\documentstyle[proceed,chicagor,smallsec,url,amssymb]{article}
\newtheorem{THEOREM}{Theorem}[section]
\newenvironment{theorem}{\begin{THEOREM} \hspace{-.85em} {\bf :} }%
                        {\end{THEOREM}}
\newtheorem{LEMMA}[THEOREM]{Lemma}
\newenvironment{lemma}{\begin{LEMMA} \hspace{-.85em} {\bf :} }%
                      {\end{LEMMA}}
\newtheorem{COROLLARY}[THEOREM]{Corollary}
\newenvironment{corollary}{\begin{COROLLARY} \hspace{-.85em} {\bf :} }%
                          {\end{COROLLARY}}
\newtheorem{PROPOSITION}[THEOREM]{Proposition}
\newenvironment{proposition}{\begin{PROPOSITION} \hspace{-.85em} {\bf :} }%
                            {\end{PROPOSITION}}
\newtheorem{DEFINITION}[THEOREM]{Definition}
\newenvironment{definition}{\begin{DEFINITION} \hspace{-.85em} {\bf :} \rm}%
                            {\end{DEFINITION}}
\newtheorem{CLAIM}[THEOREM]{Claim}
\newenvironment{claim}{\begin{CLAIM} \hspace{-.85em} {\bf :} \rm}%
                            {\end{CLAIM}}
\newtheorem{EXAMPLE}[THEOREM]{Example}
\newenvironment{example}{\begin{EXAMPLE} \hspace{-.85em} {\bf :} \rm}%
                            {\end{EXAMPLE}}
\newtheorem{REMARK}[THEOREM]{Remark}
\newenvironment{remark}{\begin{REMARK} \hspace{-.85em} {\bf :} \rm}%
                            {\end{REMARK}}

\newcommand{\thm}{\begin{theorem}}
\newcommand{\lem}{\begin{lemma}}
\newcommand{\pro}{\begin{proposition}}
\newcommand{\dfn}{\begin{definition}}
\newcommand{\rem}{\begin{remark}}
\newcommand{\xam}{\begin{example}}
\newcommand{\cor}{\begin{corollary}}
\newcommand{\prf}{\noindent{\bf Proof:} }
\newcommand{\ethm}{\end{theorem}}
\newcommand{\elem}{\end{lemma}}
\newcommand{\epro}{\end{proposition}}
\newcommand{\edfn}{\bbox\end{definition}}
\newcommand{\erem}{\bbox\end{remark}}
\newcommand{\exam}{\bbox\end{example}}
\newcommand{\ecor}{\end{corollary}}
\newcommand{\eprf}{\bbox\vspace{0.1in}}
\newcommand{\beqn}{\begin{equation}}
\newcommand{\eeqn}{\end{equation}}

\newcommand{\bbox}{\vrule height7pt width4pt depth1pt}

\newcommand{\clm}{\begin{claim}}
\newcommand{\eclm}{\end{claim}}

\newcommand{\sat}{\models}

\newcommand{\rimp}{\Rightarrow}

\newcommand{\dimp}{\Leftrightarrow}
\newcommand{\bor}{\bigvee}

\newcommand{\union}{\cup}
\newcommand{\inter}{\cap}

\renewcommand{\phi}{\varphi}

\newcommand{\A}{{\cal A}}

\newcommand{\C}{{\cal C}}

\newcommand{\K}{{\cal K}}
\newcommand{\M}{{\cal M}}
\newcommand{\N}{{\cal N}}

\newcommand{\V}{{\cal V}}

\newcommand{\X}{{\cal X}}

\newcommand{\ol}{\setlength{\itemsep}{0pt}\begin{enumerate}}
\newcommand{\eol}{\end{enumerate}\setlength{\itemsep}{-\parsep}}
\newcommand{\ul}{\setlength{\itemsep}{0pt}\begin{itemize}}
\newcommand{\dl}{\setlength{\itemsep}{0pt}\begin{description}}
\newcommand{\edl}{\end{description}\setlength{\itemsep}{-\parsep}}
\newcommand{\eul}{\end{itemize}\setlength{\itemsep}{-\parsep}}

\newcommand{\true}{\mbox{{\it true}}}

\newcommand{\commentout}[1]{}

\newcommand{\bi}{\begin{itemize}}
\newcommand{\ei}{\end{itemize}}
\newcommand{\be}{\begin{enumerate}}
\newcommand{\ee}{\end{enumerate}}

\newenvironment{oldthm}[1]{\par\noindent{\bf Theorem #1:} \em \noindent}{\par}
\newenvironment{oldlem}[1]{\par\noindent{\bf Lemma #1:} \em \noindent}{\par}
\newenvironment{oldcor}[1]{\par\noindent{\bf Corollary #1:} \em \noindent}{\par}
\newenvironment{oldpro}[1]{\par\noindent{\bf Proposition #1:} \em \noindent}{\par}
\newcommand{\othm}[1]{\begin{oldthm}{\ref{#1}}}
\newcommand{\eothm}{\end{oldthm} \medskip}
\newcommand{\olem}[1]{\begin{oldlem}{\ref{#1}}}
\newcommand{\eolem}{\end{oldlem} \medskip}
\newcommand{\ocor}[1]{\begin{oldcor}{\ref{#1}}}
\newcommand{\eocor}{\end{oldcor} \medskip}
\newcommand{\opro}[1]{\begin{oldpro}{\ref{#1}}}
\newcommand{\eopro}{\end{oldpro} \medskip}

\newcommand{\bxor}[1]{\dot{\bor}}

\newcommand{\shortv}{\commentout}
\newcommand{\fullv}[1]{#1}

\newcommand{\LKAn}{{\cal L}^{K,X,A}_n}

\newcommand{\LXAn}{{\cal L}^{X,A}_n}

\newcommand{\undefined}{\uparrow\!\!}

\renewcommand{\L}{{\cal L}}

\newcommand{\LQKXAn}{{\cal L}^{\forall,K,X,A}_n}
\newcommand{\LQKn}{{\cal L}^{\forall,K}_n}

\newcommand{\LQXAn}{{\cal L}^{\forall,X,A}_n}
\newcommand{\LKn}{{\cal L}^{K}_n}

\newtheorem{FACT}[THEOREM]{Fact}
                            {\end{FACT}}

\renewcommand{\Sigma}{S}
\newcommand{\agpp}{\mathit{agpp}}
\newcommand{\ka}{\mathit{ka}}
\newcommand{\axXn}{\mathrm{AX}^{X,A,\forall}_e}
\newcommand{\axKon}{\mathrm{AX}^{K,X,A,\forall}}
\newcommand{\axXon}{\mathrm{AX}^{X,A,\forall}}
\newcommand{\axKn}{\mathrm{AX}^{K,X,A,A^*,\forall}_e}
\newcommand{\axKnrr}{\mathrm{AX}^{K,A^*}_e}
\newcommand{\axKnr}{\mathrm{AX}^{K,A^*,\forall}_e}

\title{Reasoning About Knowledge of Unawareness Revisited}

\author{ {\bf Joseph Y. Halpern}\\
Computer Science Department\\
Cornell University\\
Ithaca, NY, 14853, U.S.A.\\
halpern@cs.cornell.edu
\And
{\bf Leandro C. R\^ego}\\
Statistics Department\\
Federal University of Pernambuco\\
Recife, PE, 50740-040, Brazil \\
leandro@de.ufpe.br
}

\begin{document}

\maketitle

\begin{abstract}
In earlier work \cite{HR05b}, we proposed a logic that extends the
Logic of General Awareness of Fagin and Halpern \citeyear{FH}
by allowing quantification over primitive propositions.  This makes it
possible to express the fact that an agent knows that there are some facts
of which he is unaware.  In that logic, it is not possible to
model an agent who is uncertain about whether he is aware of all formulas.
To overcome this problem, we keep the syntax of the earlier paper, but
allow models where, with each world, a possibly different language is
associated. We
provide a sound and complete axiomatization for this logic and show
that, under natural assumptions,
the quantifier-free fragment of the logic is
characterized by exactly the same axioms as the logic of Heifetz, Meier,
and Schipper \citeyear{HMS08b}.
\end{abstract}
\section{INTRODUCTION}
\label{intro}

Adding awareness to standard models of epistemic logic has
been
shown to be
useful in describing many situations
(see \cite{FH,HMS03} for some
examples). One of the best-known
models of awareness is due to Fagin and Halpern \citeyear{FH}
(FH from now on).
They add an awareness operator to the language, and associate with each
world in a standard possible-worlds model of knowledge a set of formulas
that each agent is aware of.  They then say that an agent
\emph{explicitly knows} a formula $\phi$ if $\phi$ is true in all worlds
that the agent considers possible (the traditional definition of
knowledge, going back to Hintikka \citeyear{Hi1}) and the agent is aware of
$\phi$.

In the economics literature, going back to the work of Modica and
Rustichini \citeyear{MR94,MR99} (MR from now on), a somewhat different
approach is taken.
A possibly different set $\L(s)$ of primitive
propositions is associated with each world $s$.  Intuitively, at world
$s$, the agent is aware only of formulas that use the primitive
propositions in $\L(s)$.   A definition of knowledge is given in this
framework, and the agent is said to be aware of $\phi$ if, by
definition, $K_i \phi \lor K_i \neg K_i \phi$ holds.  Heifetz, Meier,
and Schipper~\citeyear{HMS03,HMS08b} (HMS from now on), extend the ideas of
MR
to a multiagent setting.  This extension is
nontrivial, requiring lattices of state spaces, with projection
functions between them.  As we showed in earlier work \cite{Hal34,HR05},
the work of MR and HMS can be seen as a special case of the FH approach,
where two assumptions are made on awareness: awareness
is {\em generated by primitive propositions}, that is, an agent is aware
of a formula iff he is aware of all primitive propositions occurring in
it, and agents know what they are aware of (so that they are aware of
the same formulas in all worlds that they consider possible).
As we pointed out in \cite{HR05b} (referred to as HR from now on), if
awareness
is generated by primitive propositions, then it is
impossible for an agent to (explicitly) know that he is unaware of a
specific fact. Nevertheless,
an agent may well be aware that there are relevant facts that he is
unaware of.  For
example, primary-care physicians know that specialists are aware of things that
could improve a patient's treatment that they are not aware of; investors know
that investment fund companies may be aware of issues involving the
financial market that could result in higher profits that they are not
aware of.
It thus becomes of interest to model knowledge of lack of awareness.
HR does this by extending the syntax of the FH approach
to allow quantification, making it
possible to say that an agent knows that there exists a formula of which
the agent is unaware.
A complete axiomatization is provided for the resulting logic.
Unfortunately, the logic has a significant problem if we assume the
standard properties of knowledge
and awareness: it is impossible for an agent to be uncertain
about whether he is aware
of all formulas.
In this paper, we deal with this problem by considering the same
language as in HR (so that we can express the fact
that an agent knows that he is not aware of all formulas, using
quantification),
but using the idea of MR that there is a different language associated
with each world.  As we show,
this slight change makes it possible for an agent to
be
uncertain about whether he is aware of all formulas,
while still being aware of exactly the same formulas in all worlds he
considers possible.
We provide a natural complete axiomatization for the resulting logic.
Interestingly, knowledge in this logic acts much like explicit knowledge
in the original FH framework, if we take ``awareness of $\phi$'' to mean
$K_i(\phi \lor \neg\phi)$; intuitively, this is true if all the
primitive propositions in $\phi$ are part of the language at all worlds
that $i$ considers possible.  Under minimal assumptions,
$K_i(\phi \lor \neg \phi)$ is shown to be equivalent to $K_i \phi \lor
K_i \neg K_i \phi$: in fact,
the  quantifier-free fragment of the logic that just uses the $K_i$
operator is shown
to be characterized by exactly the same axioms
as the HMS approach, and awareness can be defined the same way.  Thus,
we can capture the essence of MR and HMS approach using simple semantics
and being able to reason about knowledge of lack of awareness.
Board and Chung~\citeyear{BC09} independently pointed
out the problem of the HR model and proposed the solution of allowing
different languages at different worlds. They also consider a model of
awareness with quantification, but they use first-order modal
logic, so their quantification is over domain elements.
Moreover, they take awareness with respect to domain elements, not
formulas; that is, agents are (un)aware of objects (i.e., domain
elements), not formulas.
They also allow different domains at different
worlds;
more precisely, they allow an agent to have a subjective view of what
the set of objects is at each world.
Sillari~\citeyear{Sil08} uses much the
same approach as Board and Chung~\citeyear{BC09}.  That is, he has a
first-order logic
of awareness, where the quantification and awareness is with respect to
domain elements, and also allows from different subjective domains at
each world.

The rest of the paper is organized as follows.
In Section~\ref{sec:awaofunawa}, we review
the
HR model of knowledge of
unawareness.  In
Section~\ref{sec:newmodel}, we present our new logic and axiomatize it in
Section~\ref{sec:axioms}.
In Section~\ref{sec:compare}, we compare our logic
with that of HMS
and discuss awareness more generally.
All proofs are left to the \shortv{full paper, which can be found at
www.cs.cornell.edu/home/halpern/papers/tark09.pdf.}
\fullv{the appendix.}

\section{THE HR MODEL}
\label{sec:awaofunawa}

In this section, we briefly review
the relevant results of \cite{HR05b}.
The syntax
of the logic is as follows: given a set $\{1, \ldots, n\}$ of
agents, formulas are formed by starting with a countable set $\Phi =
\{p, q,
\ldots\}$ of primitive propositions and a countable set  $\X$ of variables,
and then closing off under
conjunction ($\land$), negation ($\neg$), the modal operators
$K_i, A_i, X_i$, $i = 1, \ldots, n$. We also allow for
quantification over variables, so that if $\phi$ is a formula, then so is
$\forall x \phi$.
Let $\LQKXAn(\Phi,\X)$ denote this language and let
$\LKAn(\Phi)$ be the subset of formulas that do not mention quantification or variables.
As usual, we define $\phi \lor \psi$, $\phi \rimp
\psi$, and $\exists x\varphi$ as abbreviations of $\neg (\neg \phi \land \neg \psi)$,
$\neg \phi \lor \psi$, and $\neg\forall
x\neg\varphi$, respectively. The intended interpretation of
$A_i\varphi$ is ``$i$ is aware of $\varphi$''.

Essentially as in first-order logic, we can define
inductively what it means for a variable $x$ to be {\it
free} in a formula $\varphi$. Intuitively, an occurrence of a variable is free in a formula if it
is not bound by a quantifier. A formula that contains no free
variables is called a {\it sentence}.
We are ultimately interested in sentences.
If $\psi$ is a formula, let $\varphi[x/\psi]$ denote
the formula that results by replacing
all free occurrences of the variable $x$ in $\phi$ by $\psi$. (If
there is no free occurrence of $x$ in $\varphi$, then
$\varphi[x/\psi]=\varphi$.)
In quantified modal logic, the quantifiers are typically taken to range
over propositions (intuitively, sets of worlds), but this does not work
in our setting because awareness is syntactic;  when we write, for
example, $\forall x A_i x$, we essentially mean that $A_i \phi$ holds for all
\emph{formulas} $\phi$.  However, there is another subtlety.  If we
define $\forall x \phi$ to be true if $\phi[x/\psi]$ is true for \emph{all}
formulas $\psi$, then there are problems giving semantics to a formula
such as  $\phi = \forall x (x)$, since $\phi[x/\phi] = \phi$.  We avoid
these difficulties by taking the quantification to be over
quantifier-free sentences.
(See \cite{HR05b} for further discussion.)

We give semantics to
sentences in $\LQKXAn(\Phi,\X)$ in awareness
structures.
A tuple $M =(\Sigma$, $\pi$, ${\cal K}_1$, $\ldots$, ${\cal K}_n$, ${\cal
A}_1$, $\dots$, ${\cal A}_n)$ is an {\it awareness structure for $n$
agents (over $\Phi$)\/} if $\Sigma$ is a set
of worlds, $\pi: \Sigma
\times \Phi \rightarrow \{{\bf true},{\bf false}\}$ is an
interpretation that determines which primitive propositions are true
at each world, ${\cal K}_i$ is a binary relation on $\Sigma$ for
each agent $i = 1, \ldots, n$, and ${\cal A}_i$ is a function
associating a set of sentences with each world in $S$, for $i=
1,...,n$. Intuitively, if $(s,t) \in \K_i$, then agent $i$ considers
world $t$ possible at world $s$, while ${\cal A}_i(s)$ is the set of
sentences
that agent $i$ is aware of at world $s$.
We are often interested in awareness structures where the $\K_i$
relations satisfy some properties of interest, such as reflexivity,
transitivity, or the \emph{Euclidean} property (if $(s,t), (s,u) \in
\K_i$, then $(t,u) \in \K_i$).  It is well known that these properties
of the relation correspond to properties of knowledge of interest (see
Theorem~\ref{thm:awofunaaxiomswithoutK} and the following discussion).
We often abuse notation and define $\K_i(s) = \{t: (s,t) \in \K_i\}$,
thus writing $t \in \K_i(s)$ rather than $(s,t) \in \K_i$.
This notation allows us to view a binary relation $\K_i$ on $\Sigma$ as a
\emph{possibility correspondence}, that is, a function from $\Sigma$ to
$2^{\Sigma}$.  (The use of possibility correspondences is more standard
in the economics literature than binary relations, but they are clearly
essentially equivalent.)

Semantics is given to sentences in $\LQKXAn(\Phi,\X)$
by induction on the number of quantifiers, with
a
subinduction on the length of the
sentence.
Truth for primitive propositions, for $\neg$, and for
$\wedge$ is defined in the usual way.  The other cases are defined as
follows:%
\footnote{HR gives semantics to arbitrary formulas,
including formulas with free variables.  This requires using
\emph{valuations} that give meaning to free variables.  By restricting
to sentences, which is all we are ultimately interested in, we are able
to dispense with valuations here, and thus simplify the presentation of the
semantics.}
$$
\begin{array}{l}
(M,s)\sat K_i \varphi \mbox{ if }
(M,t)\sat \varphi \mbox{ for all }t
\in \K_i(s)
\\
(M,s)\sat A_i\varphi\mbox{ if }\varphi\in {\cal A}_i(s)\\
(M,s)\sat X_i\varphi\mbox{ if }(M,s)\sat A_i\varphi\mbox{ and
}(M,s)\sat K_i\varphi
\\
(M,s)\sat \forall x\varphi\mbox{ if }(M,s) \sat \phi[x/\psi], \forall\psi \in \LKAn(\Phi).
\end{array}
$$

There are two standard restrictions on agents' awareness that capture
the assumptions
typically made in the game-theoretic literature
\cite{MR99,HMS03,HMS08b}.  We describe these here in terms of the awareness
function, and then characterize them axiomatically.
\begin{itemize}
\item Awareness is {\em generated by
primitive propositions
(agpp)
\/} if, for all agents $i$, $\phi \in \A_i(s)$
iff all the primitive propositions that appear in $\phi$ are in
$\A_i(s) \inter \Phi$.
\item {\em Agents know what they are aware of
(ka)
\/} if, for all agents
$i$ and all worlds $s,t$ such that $(s,t) \in \K_i$ we have that
$\A_i(s)~=~\A_i(t)$.
\end{itemize}

For ease of exposition, we restrict in this paper to structures that
satisfy $\agpp$ and $\ka$.
If $C$ is a (possibly empty) subset of $\{r,t,e\}$,
then $\M_n^{C}(\Phi,\X)$ is the set
of all awareness structures
such that awareness satisfies $\agpp$ and $\ka$ and the possibility
correspondence is
reflexive ($r$), transitive ($t$), and Euclidean ($e$) if these
properties are in $C$.

A sentence $\varphi\in \LQKXAn(\Phi,\X)$ is said to be {\it valid} in
awareness
structure $M$, written $M \sat \phi$,
if $(M,s)\not\sat\neg\varphi$ for all $s\in S$.
(This notion is called \emph{weak validity} in \cite{HR05}.
For the semantics we are considering here, weak validity is equivalent to the
standard notion of
validity, where a formula is valid in an awareness structure if it is
true at all worlds in that structure. However, in the next section,
we modify the semantics to allow some formulas to be undefined
at some worlds; with this change, the two notions
do not coincide. As we use weak validity in the next section, we use the
same definition here for the sake of uniformity.)
A
sentence is valid in a class $\M$ of awareness structures, written $\M
\sat \phi$, if
it is valid for all awareness structures in $\M$,
that is, if $M \sat \phi$ for all $M \in \M$.

In \cite{HR05b}, we gave sound and complete axiomatizations for both the
language $\LQKXAn(\Phi,\X)$ and the language
$\LQXAn(\Phi,\X)$, which does not mention the implicit
knowledge operator $K_i$ (and the quantification is thus only over
sentences in $\LXAn(\Phi)$). The latter language is arguably more natural
(since agents do not have access to the implicit knowledge modeled by
$K_i$), but some issues become clearer when considering both.
We start by describing axioms for the language $\LQKXAn(\Phi,\X)$, and
then describe how they are modified to deal with $\LQXAn(\Phi,\X)$.
Given a formula $\phi$, let $\Phi(\phi)$ be the set of primitive
propositions in $\Phi$ that occur in $\phi$.

\begin{description}

\item[{\rm Prop.}] All substitution instances of valid formulas of
propositional logic.

\item[{\rm AGPP.}] $A_i\phi \dimp
\land_{p\in\Phi(\phi)}
A_i p.$\footnote{As usual, the empty conjunction is taken to be
the vacuously true formula $\true$, so that $A_i \phi$ is vacuously true
if no primitive
propositions occur in $\phi$.
We remark that in the conference version of HR, an apparently weaker
version of AGPP called \emph{weak generation of awareness by primitive
propositions} is used.  However, this is shown in HR to be equivalent to
AGPP if the agent is aware of at least one primitive proposition, so
AGPP is used in the final version of HR, and we use it here as well.}

\item[{\rm KA.}] $A_i \phi \rimp K_i A_i \phi$

\item[{\rm NKA.}] $\neg A_i \phi \rimp K_i \neg A_i \phi$

\item[{\rm K.}] $(K_i\varphi\land K_i(\varphi\rimp\psi))\rimp
K_i\psi$.

\item[{\rm T.}] $K_i\varphi\rimp \varphi$.

\item[{\rm 4.}] $K_i\varphi\rimp K_iK_i\varphi$.

\item[{\rm 5.}] $\neg K_i\varphi\rimp K_i\neg K_i\varphi$.

\item[{\rm A0.}] $X_i\varphi\dimp K_i\varphi\land A_i\varphi$.

\item[{\rm $1_{\forall}$.}] $\forall x\varphi\rimp\varphi[x/\psi]$
if $\psi$ is
a quantifier-free sentence.
\item[{\rm ${\rm K}_{\forall}$.}] $\forall x(\varphi\rimp\psi)\rimp (\forall
x \varphi\rimp\forall x\psi)$.

\item[{\rm ${\rm N}_{\forall}$.}] $\varphi\rimp\forall x\varphi$ if $x$ is not free
in $\varphi$.

\item[{\rm Barcan.}] $\forall xK_i\varphi\rimp K_i\forall x\varphi$.

\item[{\rm MP.}] {F}rom $\varphi$ and $\varphi\rimp\psi$ infer $\psi$
(modus ponens).

\item[{\rm Gen$_K$.}] {F}rom $\varphi$ infer $K_i \varphi$.

\item[{\rm Gen$_{\forall}$.}]
If $q$ is a primitive proposition, then from $\phi$ infer $\forall
x\varphi[q/x]$.
\end{description}

Axioms Prop, K, T, 4, 5 and inference rules MP and Gen$_K$ are standard
in epistemic logics.
A0 captures the relationship between explicit knowledge, implicit knowledge and awareness.
Axioms 1$_\forall$, K$_\forall$, N$_\forall$ and inference rules Gen$_\forall$ are standard for propositional quantification.%
\footnote{Since we gave semantics not just to sentences, but also
to formulas with free variables in \cite{HR05b}, we were able to use a
simpler version of Gen$_\forall$ that applies to arbitrary
formulas: from $\phi$ infer $\forall x \phi$.  Note that all
the other axioms and
inference rules apply without change to formulas as well as sentences.}
The Barcan axiom, which is well-known in first-order modal logic, captures the
relationship between quantification and $K_i$.
Axioms AGPP, KA, and NKA capture the properties of awareness being generated
by primitive propositions and agents knowing which formulas they are aware of.
\commentout{
\begin{description}
\item[{\rm Prop.}] All substitution instances of formulas valid
in propositional logic.

\item[{\rm AGPP.}] $A_i\phi \dimp \land_{\{p \in \Phi:\ p\  \mathrm{occurs\  in} \ \phi\}}
A_i p.$\footnote{As usual, the empty conjunction is taken to be
vacuously true, so that $A_i \phi$ is vacuously true if no primitive
propositions occur in $\phi$.}

\item[{\rm A0$_{X}$.}] $X_i\varphi\rimp A_i\varphi$.

\item[{\rm K$_{X}$.}] $(X_i\varphi\land X_i(\varphi\rimp\psi))\rimp
X_i\psi$.

\item[{\rm T$_{X}$.}] $X_i\varphi\rimp \varphi$.

\item[{\rm 4$_{X}$.}] $(X_i\varphi\land X_iA_i\phi)\rimp X_iX_i\varphi$.

\item[{\rm 5$_{X}$.}] $(\neg X_i\varphi \land A_i\varphi) \rimp
X_i\neg X_i\varphi$.

\item[{\rm Barcan$_X$.}] $\forall xX_i\varphi\rimp X_i\forall x\varphi$.

\item[{\rm $1_{\forall}$.}] $\forall x\varphi\rimp\varphi[x/\psi]$
if $\psi$ is
a quantifier-free sentence.

\item[{\rm ${\rm K}_{\forall}$.}] $\forall x(\varphi\rimp\psi)\rimp (\forall
x \varphi\rimp\forall x\psi)$.

\item[{\rm ${\rm N}_{\forall}$.}] $\varphi\rimp\forall x\varphi$ if $x$ is not free
in $\varphi$.

\item[{\rm MP.}] {F}rom $\varphi$ and $\varphi\rimp\psi$ infer $\psi$
(modus ponens).

\item[{\rm Gen$_X$.}] {F}rom $\varphi$ infer $A_i\varphi\Rightarrow X_i
\varphi$.

\item[{\rm Gen$_{\forall}$.}]
If $q$ is a primitive proposition, then from $\phi$ infer $\forall
x\varphi[q/x]$.

\end{description}
Let ${\mathbb X}_n^{\forall}$ be the axiom system consisting of all the
the axioms and inference rules above, except for T$_X$, 4$_X$, and
5$_X$.
}
Let $\axKon$ be the axiom system consisting of all the
axioms and inference rules in $\{$Prop, AGPP, KA, NKA, K, A0, 1$_\forall$, K$_\forall$,
N$_\forall$, Barcan, MP, Gen$_K$, Gen$_\forall\}$.

The language $\LQXAn$ without the modal operators $K_i$ has
an axiomatization that is similar in spirit.  Let K$_X$, T$_X$, 4$_X$, XA, and
Barcan$_X$ be
the axioms that result by replacing the $K_i$ in K, T, 4, KA, and
Barcan, respectively, by $X_i$.
Let 5$_X$ and Gen$_X$ be the axioms
that result from adding awareness to 5 and Gen$_K$:
\begin{description}

\item[{\rm 5$_{X}$.}] $(\neg X_i\varphi \land A_i\varphi) \rimp
X_i\neg X_i\varphi$.

\item[{\rm Gen$_X$.}] {F}rom $\varphi$ infer $A_i\varphi\Rightarrow X_i
\varphi$.
\end{description}
The analogue of axiom NKA written in terms of $X_i$, $\neg A_i \phi
\rimp X_i \neg A_i \phi$, is not valid.
To get completeness in models where agents know
what they are aware of, we need the following axiom,
which can be viewed as a weakening of NKA:
\begin{description}
\item[\rm{FA$_X$.}] $\neg \forall x A_i x\rimp X_i \neg \forall x A_i x$.
\end{description}

Finally, consider the following axiom that captures the relation between explicit knowledge and awareness:
\begin{description}
\item[{\rm A0$_X$.}] $X_i\phi\rimp A_i\phi$.
\end{description}
Let $\axXon$ be the axiom system consisting of all the
the axioms and inference rules in
$\{$Prop, AGPP, XA, FA$_X$, K$_X$, A0$_X$, 1$_\forall$, K$_\forall$,
N$_\forall$, Barcan$_X$, MP, Gen$_X$, Gen$_\forall\}$.
The following result shows that the semantic properties $r, t, e$ are captured by the axioms
T, 4, and 5, respectively in the language $\LQKXAn$; similarly,
these same properties are captured by T$_X$,
4$_X$, and 5$_X$ in the language $\LQXAn$.

\thm
\label{thm:awofunaaxiomswithoutK}
{\rm \cite{HR05b}}
If $\C$ (resp., $\C_X$) is a (possibly empty) subset of
$\{\rm{T}, 4, 5\}$
(resp., $\{\rm{T}_X, 4_X, 5_X\}$)
and if $C$ is the corresponding subset of
$\{r, t, e\}$
then $\axKon \union \C$
(resp., $\axXon \union \C_X$)
is a sound and
complete axiomatization of the
sentences in $\LQKXAn(\Phi,\X)$
(resp. $\LQXAn(\Phi,\X)$)
with
respect to $\M_n^{C}(\Phi,\X)$.%
\ethm

Consider the formula
$\psi=
\neg X_i \neg \forall x A_ix\land \neg X_i\forall x A_ix.$
The formula $\psi$ says that agent $i$ considers it possible
that she is aware of all formulas and also considers it possible that
she is not aware of all formulas.
It is not hard to show $\psi$ is not satisfiable in any  structure in
$\M(\Phi,\X)$, so $\neg \psi$ is
valid in awareness structures in
$\M(\Phi,\X)$,
\commentout{
Nevertheless, $\neg \psi$ is not provable in $\axXxon \union
\{AGPP,XA,5_X\}$;
this will follow from the results proved in the next section.
(Of course, it is provable in ${\bf K}_n^{\forall} \union \{AGPP,KA,5\}$,
by Theorem~\ref{thm:awofunaaxiomswithoutK}.)
Although we could possibly add further axioms to
${\bf X}_n^{\forall} \union \{5_X\}$ to get an axiom system that is
sound and complete for the language $\LQXAn(\Phi,\X)$,
we do not believe that this is such an interesting research direction.
}
It seems reasonable that an agent can be uncertain about whether there
are formulas he is unaware of.
In the next section, we show that a slight modification of the HR
approach using ideas of MR, allows this,
while still maintaining the desirable properties of the HR approach.

\section{THE NEW MODEL}
\label{sec:newmodel}

We keep the syntax of Section~\ref{sec:awaofunawa}, but, following MR,
we allow different languages to be associated with different worlds.
Define an {\em extended awareness
structure for $n$ agents (over $\Phi$)\/} to be a tuple $M =(S$, $\L$,
$\pi$, ${\cal K}_1$, $\ldots$, ${\cal K}_n,{\cal
A}_1$, $\dots$, ${\cal A}_n)$, where
$M =(S$, $\pi$, ${\cal K}_1$, $\ldots$, ${\cal K}_n,{\cal
A}_1$, $\dots$, ${\cal A}_n)$ is an awareness structure and $\L$ maps
worlds in $S$ to
nonempty
subsets of $\Phi$.
Intuitively, $\LQKXAn(\L(s),\X)$ is the
language associated with world $s$.  We require that $\A_i(s) \subseteq
\LQKXAn(\L(s),\X)$, so that
an agent can be aware only of sentences that are in the language
of the current world.
We still want to require that $\agpp$ and $\ka$; this means that if
$(s,t) \in \K_i$, then $\A_i(s) \subseteq \LQKXAn(\L(t),\X)$.  But
$\L(t)$ may well include primitive propositions that the agent is not
aware of at $s$.
It may at first seem strange that an agent
considers possible a
world whose language includes formulas of which he is not aware.  (Note
that, in general, this happens in the HR approach too, even though there
we require that $\L(s) = \L(t)$.)  But, in the context of knowledge of
lack awareness, there is an easy explanation for this: the fact that
$\A_i(s)$ is a strict subset of the sentences in $\LQKXAn(\L(t),\X)$ is just our way of
modeling that the agent considers it possible that there are formulas of
which he is unaware; he can even ``name''
or ``label''
these formulas, although he
may not understand what the names refer to.  If the agent considers
possible a world $t$ where $\A_i(s)$ consists of every sentence in
$\LQKXAn(\L(t),\X)$, then the agent
considers it possible that he is aware of all formulas.  The formula
$\psi$ in Section~\ref{sec:awaofunawa} is satisfied at a world $s$ where
\commentout{
agent $i$ considers possible a world $t_1$ such that $\A_i(s) =
\LQKXAn(\L(t_1),\X)$ and a world $t_2$ such that $\A_i(s)$ is a strict
subset of
$\LQKXAn(\L(t_2),\X)$.  Note that we can also describe worlds where agent
}
agent $i$ considers possible a world $t_1$ such that
$\A_i(s)$ consists of all sentences in  $\LQKXAn(\L(t_1),\X)$ and a
world $t_2$ such that $\A_i(s)$ does not
contain some sentence in
$\LQKXAn(\L(t_2),\X)$.  Note that we can also describe worlds where agent
1 considers it possible that agents 2 and 3 are aware of the same
formulas, although both are aware of formulas that he (1) is not aware
of, and other more complicated relationships between the awareness of
agents.
See Section~\ref{sec:compare} for further discussion of awareness of
unawareness in this setting.

\commentout{
Depending on the
interpretation of awareness one has in mind,
other restrictions on
${\cal A}_i$ may apply. We discuss some interesting restrictions later
on. To give semantics to primitive variables we use syntactic
valuations. A {\em syntactic valuation} is a function $\V(x,s)\in
\LKAn(\L(s))$, which assigns to each variable $x$ and world $s$ a
sentence in
$\LKAn(\L(s))$. We write $\V \sim_{x,s} \V'$ if $\V(y,s) = \V'(y,s)$ for all variables $y \ne x$.

We write $(M,s,\V)\sat \varphi$ if $\varphi$ is true at world $s$ in
the awareness structure $M$ given the syntactic valuation $\V$.
}

The
truth relation is defined
for formulas in $\LQKXAn(\Phi,\X)$
just as in Section~\ref{sec:awaofunawa}, except that for a formula
$\phi$ to be true at a world
$s$, we also require that $\phi \in \LQKXAn(\L(s),\X)$, so we just add
this condition everywhere.  Thus, for
example,
\begin{itemize}
\item $(M,s)\sat p$ if $p\in\L(s)$ and $\pi(s,p)= {\bf true}$;
\item $(M,s)\sat \neg \phi$ if $\phi\in \LQKXAn(\L(s),\X)$ and
$(M,s)\not\sat \phi$.
\item $(M,s)\sat \forall x\varphi$ if $\varphi\in \LQKXAn(\L(s),\X)$ and \\
$(M,s) \sat \phi[x/\psi]$ for
all $\psi \in \LKAn(\L(s))$.
\end{itemize}
We leave it to the reader to make the obvious changes to the remaining
clauses.

\commentout{
A formula $\varphi\in \LQKXAn(\Phi,\X)$ is said to be {\it valid} in
extended awareness
structure $M$,
written $M \sat \phi$,
if $(M,s,\V)\not\sat\neg\varphi$ for all $s\in S$ and $\V$. A
formula is valid in a class $\N$ of extended awareness structures, written $\N
\sat \phi$, if
it is valid for all extended awareness structures in ${\cal N}$,
that is, if $N \sat \phi$ for all $N \in {\cal N}$.
}

If $C$ be a (possibly empty) subset of $\{r,t,e\}$,
$\N_n^{C}(\Phi,\X)$ be the set
of all extended awareness structures
such that awareness satisfies $\agpp$ and $\ka$ and the possibility
correspondence is
reflexive, transitive, and Euclidean if these
properties are in $C$.
We say that a formula $\phi$ is \emph{valid in a class $\N$ of extended
awareness
structures} if, for all extended awareness structures $M\in \N$ and worlds $s$
such that $\Phi(\phi) \subseteq \L(s)$, $(M,s) \sat \phi$.  (This is
essentially the notion of weak validity defined in \cite{HR05}.)

\section{AXIOMATIZATION}
\label{sec:axioms}

In this section, we provide a sound and complete axiomatization of
the logics described in the previous section.
It turns out to be easier to start with the language $\LQXAn(\Phi,\X)$.
All the axioms and inference rules of $\axXon$ continue to be sound in
extended
awareness structures, except for Barcan$_X$ and FA$_X$.
In a world $s$ where $\L(s) = p$ and agent 1 is aware of $p$, it is easy
to see that $\forall x X_i A_i x$ holds.  But if agent 1 considers
possible a world $t$ such that $\L(t) = \{p,q\}$, it is easy to see that
$X_i \forall x A_i x$ does not hold at $s$.
Similarly, if in world $t$,
agent 1 considers $s$ possible, then $\neg \forall x A_i x$ holds at
$t$, but $X_i \neg \forall x A_i x$ does not.  Thus, Barcan$_X$ does not
hold at $s$, and FA$_X$ does not hold at $t$.  We instead use the
following variants of Barcan$_X$ and FA$_X$, which are sound in this
framework:
\begin{description}
\item[{\rm Barcan$^*_X$.}] $(A_i  (\forall x \phi) \land \forall x (A_i
x \rimp X_i \phi)) \rimp X_i (\forall x A_i x \rimp \forall x \phi)$.
\item[{\rm FA$^*_X$.}] $\forall x \neg  A_i x \rimp X_i \forall x \neg
A_i x$.
\end{description}
Let $\axXn$ be the result of replacing FA$_X$ and Barcan$_X$ in $\axXon$
by FA$_X^*$ and Barcan$_X^*$ (the $e$ here stands for
``extended'').
\commentout{

\subsection{AN AXIOMATIZATION FOR THE LANGUAGE $\LQXAn(\Phi,\X)$}

\commentout{
Finally, let $A_i^*(\phi)$ be an
abbreviation for the formula $\land_{p\in \Phi(\phi)}K_i(p\lor\neg
p)$.
As usual, we take the empty conjunction to be $\true$.
Intuitively, the formula $A_i^*(\phi)$ captures the property
that $\phi$ is defined at all worlds considered possible by agent
$i$.
}

Consider the following axioms:
\begin{description}

\item[{\rm Barcan$^*_X$.}] $(A_i  (\forall x \phi) \land \forall x (A_i x \rimp X_i \phi)) \rimp X_i (\forall x A_i x \rimp \forall x \phi)$.

\item[{\rm EA.}] $\forall x \neg  A_i x \rimp X_i \forall x \neg A_i x$.
\end{description}

Because we allow a different language at each world, it is easy to see
that Barcan$_X$ is not sound under the new semantics.   In a world $s$ where
the language involves just one primitive proposition, say $p$, and agent
1 aware of $p$ (and hence all sentences in $\LQKXAn(\L(s),X)$), then
if agent 1 knows what he is aware of, then
sentence $\forall x X_1 (A_1(x))$ holds.  However, $X_1 \forall x
A_1(x)$
may not hold if agent 1 considers possible a world where he is
not aware of all formulas.  Barcan$^*_X$ is a variant of Barcan$_X$ that
is sound in models where awareness is generated by primitive
propositions and agents know what they are aware of.
Axiom AGPP captures the property $\agpp$.
It is easy to see that KA and NKA capture the property $\ka$
using $K_i$.
XA is the analogue of KA
expressed using $X_i$; it is easily seen to be valid
in structures that satisfy $\ka$.  The analogue of NKA expressed using
$X_i$, $\neg A_i \phi\rimp X_i \neg
A_i\phi$, is not valid.  Axiom
FA$_X^*$
is a weaker version that is sufficient to completely capture the
property $\ka$ in this language.%
Let $\axXn$ consist of all the
axioms and inference
rules in $\{$Prop, AGPP, XA, K$_X$, A0$_X$, 1$_\forall$, K$_\forall$, N$_\forall$,
Barcan$^*_X$, EA, MP, Gen$_X$, Gen$_\forall\}$.
}

\thm
\label{thm:compwithoutK}
If $\C_X$ is a (possibly empty) subset of
$\{\rm{T}_X, 4_X, 5_X\}$ and $C$ is the corresponding subset of $\{r,t,
e\}$, then $\axXn \union \C_X$ is a sound and
complete axiomatization of the language $\LQXAn(\Phi,\X)$ with
respect to $\N_n^{C}(\Phi,\X)$.
\ethm

The completeness proof is similar in spirit to that of HR, with some
additional complications arising from the interaction between
quantification and the fact  that different
languages  are associated with different worlds.  What is surprisingly
difficult in this case is soundness, specifically, for MP.  For suppose
that $M$ is a structure in $\N_n(\Phi,\X)$ such that neither $\neg \phi$ nor
$\neg(\phi \rimp \psi)$ are true at any world in $M$.  We want to show
that $\neg \psi$ is not true at any world in $M$.  This is easy to show if
$\Phi(\psi) \subset \Phi(\phi)$.  For if $s$ is a world such that
$\Phi(\psi) \subseteq \L(s)$, it must be the case that both $\phi$ and
$\phi \rimp \psi$ are true at $s$, and hence so is
$\psi$. However, if
$\phi$ has some primitive propositions that are not in $\psi$, it is a
priori possible that $\neg \psi$ holds at a world where neither $\phi$
nor $\phi \rimp \psi$ is defined.  Indeed, this can happen if $\Phi$ is
finite.  For example, if $\Phi = \{p,q\}$, then it is easy to construct
a structure $M \in \N_n(\Phi,X)$ where both $A_ip \land A_i q$ and $(A_i
p \land A_i q) \rimp \forall x A_i x$ are never false, but $\forall x
A_i x$ is false at some world in $M$.  As we show, this cannot happen if
$\Phi$ is infinite.  This in turn
involves proving a general substitution property: if $\phi$ is valid
and $\psi$ is a quantifier-free sentence,
then $\phi[q/\psi]$ is valid.  (We remark that the substitution property
also fails if $\Phi$ is finite.) See the \fullv{appendix} \shortv{full
paper} for details.
\shortv{Proofs for all other results stated in this abstract can also be
found in the full paper.}

Using different languages has a greater impact on the axioms for $K_i$
than it does for $X_i$.  For example, as we would
expect, Barcan does not hold, for
essentially the same reason that Barcan$_X$ does not hold.  More
interestingly, NKA, 5, and Gen$_K$ do not hold either.  For example, if
$\neg K_i p$ is true at a world $s$ because
$p \notin \L(t)$ for some world $t$
that $i$ considers possible at $s$, then $K_i \neg K_i p$ will not hold
at $s$, even if the $\K_i$ relation is an equivalence relation.
Indeed, the properties of $K_i$ in this framework become quite close to
the properties of the explicit knowledge operator $X_i$ in the original
FH framework, provided we define the appropriate variant of awareness.

Let $A_i^*(\phi)$ be an
abbreviation for the formula
$K_i(\phi \lor \neg \phi)$.
Intuitively, the formula $A_i^*(\phi)$ captures the property
that $\phi$ is defined at all worlds considered possible by agent
$i$.
Let AGPP$^*$, XA$^*$, A0$^*$, 5$^*$, Barcan$^*$, FA$^*$, and Gen$^*$ be the result of
replacing
$X_i$ by $K_i$ and
$A_i$ by $A_i^*$ in AGPP,  XA, A0$_X$, 5$_X$, Barcan$^*_X$, FA$^*_X$, and Gen$_X$,
respectively.  It is easy to see that AGPP$^*$, A0$^*$, and Gen$^*$ are
valid in
extended awareness structures; XA$^*$, 5$^*$, Barcan$^*$, and FA$^*$ are not.
For example, suppose that $p$ is
defined in all worlds that agent $i$ considers possible at $s$, so that
$A_i^*p$ holds at $s$.  If there is some world $t$ that agent $i$
considers possible at $s$ and a world $u$ that agent $i$ considers
possible at $t$ where $p$ is not defined, then $A_i^*p$ does not hold at
$t$, so $K_i A_i^*p$ does not hold at $s$.  It is easy to show that
XA$^*$ holds if the $\K_i$ relation is transitive.
Similar arguments show that 5$^*$, Barcan$^*$, and FA$^*$ do not hold in
general, but are valid if $\K_i$ is
Euclidean and (in the case of Barcan$^*$ and FA$^*$) reflexive.
We summarize these observations in the following proposition:
\pro\label{pro:soundness}
\begin{itemize}
\item[(a)] XA$^*$ is valid in $\N_n^{t}(\Phi,\X)$.
\item[(b)] Barcan$^*$ is valid in $\N_n^{r,e}(\Phi,\X)$.
\item[(c)] FA$^*$ is valid in $\N_n^{r,e}(\Phi,\X)$.
\item[(d)] 5$^*$ is valid in $\N_n^{e}(\Phi,\X)$.
\end{itemize}
\epro

In light of Proposition~\ref{pro:soundness}, for ease of exposition, we
restrict attention for the rest of this section to structures in
$\N_n^{r,t,e}(\Phi,\X)$. Assuming that the possibility relation is
an equivalence relation is standard when modeling knowledge
in any case.
\commentout{
We get axioms for the language $\LQKXAn(\Phi,\X)$ by replacing
$X_i$ and $A_i$ by $K_i$ and $A_i^*$,
in the axioms and inference rules of ${\mathbb X}_n^{\forall}$,
except that the analogue of axiom XA is not needed here;
we also need one more standard axiom (A0) to relate $K_i$, $X_i$, and
Thus, let A0$^*$, 4$^*$, 5$^*$, Gen$^*$, Barcan$^*$, and
EA$^*$ be the axioms and inference rules that result
from replacing
$X_i$ and $A_i$ by $K_i$ and $A_i^*$ in the axioms A0$_X$, 4$_X$, 5$_X$, Gen$_X$, Barcan$^*_X$, and
EA, respectively.
}
Let $\axKn$ be the result of replacing Gen$_K$ and Barcan in $\axKon$
by  Gen$^*$ and Barcan$^*$, respectively, and
adding the axioms AGPP$^*$, A0$^*$, and FA$^*$ for reasoning
about $A_i^*$.  (We do not need the axiom XA$^*$; it follows from
4 in transitive structures.)
Let $\axKnr$ consist of the axioms in $\axKn$ except for those that mention
$X_i$ or $A_i$; that is, $\axKnr = \axKn - \{$AGPP, KA, NKA, A0$\}$.
Note that $\axKnr$ is the result of replacing $X_i$ by $K_i$ and $A_i$
by $A_i^*$ in $\axXn$ (except that the analogue of XA is not needed).
Finally, let $\axKnrr$ consist of the axioms and rules in $\axKnr$
except for the ones that mention quantification; that is, $\axKnrr =
\{$Prop, AGPP$^*$, K, Gen$^*$, A0$^*\}$.  We use $\axKnrr$ to compare
our results to those of HMS.

\commentout{
The following 3 results show that Barcan$^*$ and EA$^*$ are valid only
in
structures where the $\K_i$ relation is reflexive and Euclidean, while
$A_i^*\phi \rimp
K_i A_i^*\phi$, the analogue of axiom XA,
is valid only in structures where the $\K_i$ relation is
transitive.

\thm
{\bf (Barcan$^*$)}
\label{thm:WB*}
For every formula $\phi\in\LQKXAn(\Phi,\X)$, $(A_i^*  (\forall x \phi) \land \forall x (A_i^* x \rimp K_i \phi)) \rimp K_i (\forall x A_i^* x \rimp \forall x \phi)$ is valid
in $\N_n^{r,e}(\Phi,\X)$.
\ethm

\prf
See Appendix~\ref{sec:WB*}.
\eprf

\thm
{\bf (EA$^*$)}
\label{thm:EA*}
The formula $\forall x\neg A_i^* x\rimp K_i \forall x \neg A_i^* x$ is valid
in $\N_n^{r,e}(\Phi,\X)$.
\ethm

\prf
See Appendix~\ref{sec:EA*}.
\eprf

\thm
\label{thm:KA*}
For every formula $\phi\in\LQKXAn(\Phi,\X)$, $A_i^*\phi \rimp K_i A_i^* \phi$ is valid
in $\N_n^{t}(\Phi,\X)$.
\ethm

\prf
See Appendix~\ref{sec:KA*}.
\eprf

Thus, for ease of exposition, we consider
in this section
only
structures where the $\K_i$ relation is reflexive, Euclidean, and
transitive, that is, an equivalence relation; this is essentially the
case considered by HMS as well.

\thm
\label{thm:compwithK}
${\mathbb K}_n^{\forall} \union \{\rm{T}, 4^*, 5^*\}$ is a sound and
complete axiomatization of the sentences in $\LQKXAn(\Phi,\X)$ with
respect to $\N_n^{r, e, t}(\Phi,\X)$.
\ethm

\prf
Identical to the proof of Theorem~\ref{thm:compwithoutK}, except that
$X_i$ and $A_i$ are replaced by $K_i$ and $A_i^*$, respectively, and in
Lemma~\ref{LemmaA5X}, another step is needed in the induction to deal with
$X_i$ that uses the extra axiom A0 in the standard way.
\eprf

\subsection{AXIOMATIZATIONS FOR THE LANGUAGES $\LKAn(\Phi)$ AND $\LKn(\Phi)$}
\label{sec:axwithoutquant}

HMS considered the language $\LKn(\Phi)$, without quantification and
without the $A_i$ and $X_i$ operators.  They then define $A_i \phi$ to
be an abbreviation of $K_i \phi \lor K_i \neg K_i\phi$.
Let ${\mathbb K}_n$ be the axiom system that consists of all axioms and
inference rules in ${\mathbb K}_n^{\forall}$ that do not mention
quantifiers, that is, ${\mathbb K}_n$ consists of the
axioms and inference rules in $\{$Prop, AGPP, KA, NKA, A0, A0$^*$, K, MP, Gen$^*\}$.

\begin{theorem} \label{thm:compwithoutquant} Let $\C$ be a
(possibly empty) subset of $\{\rm{T}, 4^*, 5^*\}$ and let $C$ be the corresponding subset
of $\{r, t, e\}$. Then ${\mathbb K}_n \union
\C$
(resp. ${\mathbb K}_n - \{\rm{A0}, \rm{AGPP}, \rm{KA}, \rm{NKA}\} \union \C$
is a sound and complete axiomatization of the language
$\LKAn(\Phi)$
(resp. $\LKn(\Phi)$)
with respect to
$\N_n^{C}(\Phi)$.
\end{theorem}

}
\bigskip
\bigskip

\thm\label{thm:compwithK}
\begin{itemize}
\item[(a)] $\axKn\union\{\rm{T},4,5^*\}$ is a sound and complete axiomatization of
the sentences in $\LQKXAn(\Phi,\X)$ with respect to $\N_n^{r,e,
t}(\Phi,\X)$.
\item[(b)] $\axKnr\union\{\rm{T},4,5^*\}$ is a sound and complete axiomatization of
the sentences in $\LQKn(\Phi,\X)$ with respect to $\N_n^{r,t,e}(\Phi,\X)$.
\item[(c)] $\axKnrr\union\{\rm{T},4,5^*\}$ is a sound and complete axiomatization of
$\LKn(\Phi)$ with respect to $\N_n^{r,t,e}(\Phi)$.
\end{itemize}
\ethm
Since, as we observed above, $\axKnr$ is essentially the result of
replacing $X_i$ by $K_i$ and $A_i$
by $A_i^*$ in $\axXn$,
Theorem~\ref{thm:compwithK}(b) makes precise the sense in which $K_i$ acts
like $X_i$ with respect to $A_i^*$.

\section{DISCUSSION}
\label{sec:compare}

\commentout{
In this section,
we compare
our approach to that of HMS. Since HMS do not consider
quantification, and define awareness in terms of knowledge, we do the
comparison in terms of the language
$\LKn(\Phi)$.
}

Just as in our framework, in the HMS and MR approach,
a (propositional) language is associated
with each world.
However, HMS and MR define awareness of $\phi$ as an abbreviation of
$K_i\phi \lor K_i\neg K_i\phi$.
In order to compare our
approach to that of HMS and MR, we first compare the definitions of
awareness.
Let $A_i'\phi$ be an abbreviation for the formula $K_i\phi \lor K_i\neg
K_i\phi$.
The
following result says that for extended awareness structures that are
Euclidean,
$A_i^*\phi$ is equivalent to $A_i'\phi$.

\begin{sloppypar}
\pro
\label{prop:equivdef}
If $M =(S,\L,\pi,{\cal K}_1,...,{\cal K}_n,$ ${\cal A}_1,\dots,{\cal
A}_n)$ is a Euclidean extended awareness structure, then
for all $s\in S$ and all sentences $\phi\in\LQKXAn(\Phi,\X)$,
$$(M,s)\sat A_i^*\phi \dimp A_i'\phi.$$
\epro
\end{sloppypar}

\prf
Suppose that $(M,s)\sat K_i(\phi\lor\neg \phi) \land
\neg K_i \phi$.  It follows that $\Phi(\phi)\subseteq \L(s)$,
$\Phi(\phi)\subseteq \L(t)$ for all $t$ such that $(s,t) \in \K_i$, and
that
there exists a world $t$ such that $(s,t)\in\K_i$ and
$(M,t)\sat \neg\phi$. Let $u$ be an arbitrary world such that $(s,u)\in
\K_i$. Since $\K_i$ is Euclidean, it follows that $(u,t)\in\K_i$. Thus,
$(M,u)\sat \neg K_i\phi$, so $(M,s)\sat K_i\neg K_i\phi$.   It follows that
$(M,s) \sat A_i' \phi$, as desired.

For the converse, suppose that $(M,s)\sat A_i'\phi$. If either
$(M,s)\sat K_i\phi$ or $(M,s)\sat K_i \neg K_i\phi$, then
$\Phi(\phi)\subseteq \L(s)$, and if $(s,t)\in\K_i$, we have that
$\Phi(\phi)\subseteq \L(t)$. Therefore, $(M,s)\sat A_i^*\phi$.
\eprf

\commentout{
Let $\A_i^*(s)$ be the set of sentences that are defined
at all worlds considered possible by agent $i$ in world $s$. Formally,
if $\phi$ is a sentence, then $\phi \in \A_i^*(s)$ iff $(M,s) \sat
A_i \phi$. Assuming that agents know what they are
aware of, we have
that if $(s,t) \in \K_i$, then $\A_i(s) = \A_i(t)$. Thus, it
follows that $\A_i(s)\subseteq \A_i^*(s)$. For if $\phi \in \A_i(s)$,
then
$\phi$ must be in $\L(t)$ for all $t$ such that $(s,t) \in \K_i$, so
$(M,s) \sat A_i^*(\phi)$.

We get the opposite inclusion by assuming the following natural
connection between an agent's awareness function and the language in the
worlds that he considers possible:
Indeed, we get the opposite inclusion under the following weaker assumption:
\begin{itemize}
\item {\bf LA:} If $p \notin \A_i(s)$, then $p \notin \L(t)$ for some $t$
such that $(s,t) \in \K_i$.
\end{itemize}
It is also immediate that in models that satisfy {\bf LA} (and $\agpp$),
$\A_i(s) \subseteq \A_i^*(s)$ for all agents $i$ and worlds $s$.
Thus, under minimal assumptions, $\A_i^*(s) = \A_i(s)$.
}
In \cite{HR05}, we showed that $\axKnrr \union \{\mathrm{T},4,5^*\}$ provides a sound and
complete axiomatization of the structures used by HMS where the
possibility relations are Euclidean, transitive, and reflexive%
, with one difference: $A_i'$ is used for awareness
instead of $A_i^*$.
\commentout{
One of them is ${\bf U_n}$, which contains $\{$Prop, T, 4, MP$\}$ and the following list, where $A_i\phi$ is an abbreviation for $K_i\phi\lor K_i\neg K_i\phi$, given by:
\begin{description}
\item[{\rm M.}] $K_i(\phi \land \psi) \rimp K_i\phi \land K_i\psi$.
\item[{\rm C.}] $K_i\phi \land K_i\psi \rimp K_i(\phi \land \psi)$.
\item[{\rm A.}] $A_i\phi \dimp A_i\neg\phi$.
\item[{\rm AM.}] $A_i(\phi \land \psi) \rimp A_i\phi \land A_i\psi$.
\item[{\rm N.}] $K_i\top$.
\item[{\rm REsa.}] From $\phi \dimp \psi$ infer $K_i\phi \dimp K_i\psi$, where $\phi$ and $\psi$ contain exactly the same primitive propositions.
\item[{\rm AK.}] $A_iK_j\phi\dimp A_i\phi$.
\end{description}

The other one is ${\bf S5_n^K}$, which contains $\{$Prop, T, K, 4, MP, AGPP$\}$ and the following list, where $A_i\phi$ is an abbreviation for $K_i\phi\lor K_i\neg K_i\phi$, given by:
\begin{description}
\item[{\rm 5$'$.}] $(\neg K_i\phi \land A_i\phi) \rimp K_i\neg K_i\phi$.
\item[{\rm Gen$'$.}] From $\phi$ infer $A_i\phi\rimp K_i\phi$.
\item[{\rm D1$'$.}] $K_i\phi \rimp A_i\phi$.
\item[{\rm KA.}] $A_i\phi \rimp K_iA_i\phi$.
\end{description}
}
\commentout{
One of the axiomatizations is $({\mathbb K}_n -\{$A0,
AGPP, KA, NKA$\}) \union \{\rm{T}, 4^*, 5^*\}$, which
Theorem~\ref{thm:compwithoutquant}  shows is sound and complete axiomatization for
$\LKn(\Phi)$ with respect to $\N_n^{r,t,e}$, with one difference:
$A_i'$ is used rather than $A_i^*$.
}
However, by Proposition~\ref{prop:equivdef}, in $\N_n^{e}$, $A_i^*$
and $A_i'$ are equivalent.
Thus, for the class of structures of most interest, we are able to get
all the properties of the HMS approach; moreover, we can extend to allow
for reasoning about knowledge of unawareness.  It is not clear how to
capture knowledge of unawareness directly in the HMS approach.

It remains to consider the relationship between $A_i$ and $A_i^*$.
Let $\A_i^*(s)$ be the set of sentences that are defined
at all worlds considered possible by agent $i$ in world $s$; that is,
$\phi \in \A_i^*(s)$ iff $(M,s) \sat
A_i^* \phi$. Assuming that agents know what they are
aware of, we have
that if $(s,t) \in \K_i$, then $\A_i(s) = \A_i(t)$. Thus, it
follows that $\A_i(s)\subseteq \A_i^*(s)$. For if $\phi \in \A_i(s)$,
then
$\Phi(\phi)\subseteq\L(t)$ for all $t$ such that $(s,t) \in \K_i$, so
$(M,s) \sat A_i^*(\phi)$.

We get the opposite inclusion by assuming the following natural
connection between an agent's awareness function and the language in the
worlds that he considers possible:
\begin{itemize}
\item {\bf LA:} If $p \notin \A_i(s)$, then $p \notin \L(t)$ for some $t$
such that $(s,t) \in \K_i$.
\end{itemize}
It is immediate that in models that satisfy {\bf LA} (and
$\agpp$),
$\A_i(s) \supseteq \A_i^*(s)$ for all agents $i$ and worlds $s$.
Thus, under minimal assumptions, $\A_i^*(s) = \A_i(s)$.

The bottom line here is that under the standard assumptions in the
economics literature, together with the minimal assumption {\bf LA}, all
the notions of awareness coincide.  We do not need to consider a
syntactic notion of awareness at all.  However, as pointed out by FH,
there are other notions of awareness that may be relevant; in
particular, a more computational notion of awareness is of interest.
For such a notion, an axiom such as AGPP does not seem appropriate.
We leave the problem of finding axioms that characterize a more
computational notion of awareness in this framework
to future work.

We conclude with some comments on awareness and language.
If we think of propositions $p \in \L(t) - \A_i(s)$ as just being
labels
or names
for concepts that agent $i$ is not aware of but $i$
understands other agents might be aware of, {\bf LA} is just saying that
$i$ should not use the same
label
in all worlds that he
considers possible.
It is important that an agent can use different labels for formulas that
he is unaware of.  A world where an agent is unaware of two primitive
propositions is different from a world where an agent is unaware of only
one primitive proposition.  For example, to express the fact that in
world $s$ agent
agent 1 considers it possible that (1) there is a formula that he is
unaware that agent 2 is aware of and (2) there is a formula
that both he and agent 2 are unaware of that agent 3 is aware of, agent
1 needs to consider possible a world $t$ with at least two primitive
propositions  in $\L(t) - \A_1(s)$.
Needless to say, reasoning about such lack of awareness might be
critical in a decision-theoretic context.

The fact that the primitive propositions that an agent is not aware of
are simply labels means that switching the labels does not affect what
the agent knows or believes.  More precisely, given a model $M =
(\Sigma, \L,\K_1, \ldots, \K_n, \A_1, \ldots, \A_n, \pi)$, let $M'$ be
identical to $M$ except that the roles of the primitive propositions $p$
and $p'$ are interchanged.  More formally, $M' = (\Sigma, \L', \K_1,
\ldots, \K_n, \A_1', \ldots, \A_n',\pi')$, where, for all worlds $s \in
\Sigma$, we have
\begin{itemize}
\item $\L(s) - \{p,p'\} = \L'(s) - \{p,p'\}$;
\item $p \in \L'(s)$ iff $p' \in \L(s)$, and $p' \in \L'(s)$ iff $p \in
\L(s)$;
\item $\pi(s,q) = \pi'(s,q)$ for all $q \in \L(s) - \{p,p'\}$;
\item if $p \in
\L(s)$, then $\pi(s,p) = \pi'(s,p')$, and if $p' \in \L(s)$, then $\pi(s,p')
= \pi'(s,p)$;
\item if $\phi$ is a formula that mentions neither $p$ nor $p'$, then
$\phi \in \A_i(s)$ iff $\phi \in \A_i'(s)$;
\item for any formula
$\phi$ that mentions either $p$ or $p'$, $\phi \in \A_i(s)$ iff $\phi[p
\leftrightarrow p'] \in \A_i'(s)$, where $\phi[p \leftrightarrow p']$ is the result of
replacing all occurrences of $p$ in $\phi$ by $p'$ and all occurrences
of $p'$ by $p$.
\end{itemize}
It is easy to see that for all worlds $s$,
$(M,s) \sat \phi$ iff $(M',s) \sat \phi[p
\leftrightarrow p']$.
In particular, this means that if neither $p$ nor $p'$
is in $\L(s)$, then for all formulas, $(M,s) \sat \phi$ iff $(M',s) \sat
\phi$.  Thus, switching labels of propositions that are not in $\L(s)$
has no impact on what is true at $s$.

We remark that the use of labels here is similar in spirit to our use of
\emph{virtual moves} in \cite{HR06} to model moves that a player is
aware that he is unaware of.
Although switching labels of propositions that are not in $\L(s)$
has no impact on what is true at $s$, changing the truth value of a
primitive proposition that an agent is not aware  
at $s$ may have some impact on what the agent explicitly knows at $s$.
Note that we allow agents to have some partial information
about formulas that they are unaware of. We certainly want to allow
agent 1 to know that there is a formula that agent 2 is aware of that he
(agent 1) is unaware of; indeed, capturing a situation like this was one
of our primary motivations for introducing knowledge of lack of
awareness.  But we also want to allow agent 1 to know that
agent 2 is not only aware of the formula, but knows that it is true;
that is, we want $X_1(\exists x (\neg A_1 (x) \land K_2(x)))$ to be
consistent.  There may come a point when an agent has so much
partial information about a formula he is unaware of that, although he
cannot talk about it
explicitly in his language, he can describe it sufficiently well to
communicate about it.  When this happens in natural language, people
will come up with a name for a concept and add it to their language.
We have not addressed the dynamics of language change here, but we
believe that this is a topic that deserves further research.

\commentout{
\section{Conclusion}
\label{sec:conc}
We have further advanced the problem of providing a logic to model agents who are able to reason
about their lack of  awareness. We essentially extended the class of
models considered by Halpern and R\^ego~\citeyear{HR05b}, by allowing
that the language varies depending on the world. Allowing for this extra
flexibility, gives us a way to
model an
agent
that considers
possible both that he is aware of
all formulas and that he is not aware of all formulas. We have also
provided a complete
axiomatization for this new logic
and some of its fragments. We showed that the fragment of the logic
which only mentions implicit knowledge operators is characterized by
exactly the same axioms as the logic of HMS.

Although our focus here has been on questions of logic (soundness and completeness), as we emphasized in our earlier work reasoning about awareness and
knowledge of unawareness is of great relevance to game theory. As we
pointed out in our previous work, we believe that it should be possible
to fruitfully combine ideas
from these logics with game-theoretic models.
However, to do so, we need to extend the logic
to capture probability as well
as awareness and knowledge, which we conjecture that can be done along the lines of the work in \cite{FH3,FHM}.
}

\fullv{

\appendix

\section{PROOFS}
We first prove Theorem~\ref{thm:compwithoutK}.  As we said in the main
text, proving soundness turns out to be nontrivial, so we being by
showing that MP, Barcan$^*_X$, and Gen$_\forall$ are sound.  (Soundness
of the remaining axioms is straightforward.  For MP, we need some
preliminary lemmas.

\lem
\label{lem:satiswithoutq}
If $\phi$ is a
sentence in $\LQKXAn(\Phi,\X)$
that does not mention $q$ and is satisfiable
in $\N_n(\Phi,\X)$, then it is satisfiable in an extended awareness structure
$M=(S,\L(s),\pi,\K_1,\ldots,\K_n,\A_1,\ldots,\A_n)
\in \N_n(\Phi,\X)$ such that $q\notin \L(s)$ for every $s\in S$.
\elem
\prf
Let $\tau:\Phi\rightarrow\Phi$ be a
1-1 function.
For a
sentence
$\psi$, let $\tau(\psi)$ be
the result of replacing every primitive proposition $q$ in $\psi$ by
$\tau(q)$. Given an extended awareness structure
$M^\tau(S,\L(s),\pi,\K_1,\ldots,\K_n,\A_1,\ldots,\A_n)$,  let
$M'=(S,\L^\tau(s),\pi^\tau,\K_1,\ldots,\K_n,\A^\tau_1,\ldots,\A^\tau_n)$
be the extended
awareness structure that results from ``translating'' $M$ by
$\tau$; formally:
$\L'(s)=\{\tau(p):p\in \L(s)\}$, $\pi'(s,\tau(p))=\pi(s,p)$, and
$\A'_i(s)=\{\tau(\psi):\psi\in\A_i(s)\}$.
We now prove that $(M,s)\sat \psi$ iff
$(M^\tau,s)\sat \tau(\psi)$
by induction in the structure of $\psi$.
All the cases are straightforward and left to the reader except the case
$\psi$ has the form $\forall x \psi'$.
\commentout{
\begin{itemize}
\item If $\psi$ is a primitive proposition, then the result follows from
the definition of $\pi'$.
\item Suppose that $\psi=\neg\psi'$. Then, $(M,s)\sat \psi$ iff
$\psi\in\LQKXAn(\L(s),\X)$ and $(M,s)\not\sat \psi'$. Since by
definition, $\psi\in\LQKXAn(\L(s),\X)$ iff
$\tau(\psi)\in\LQKXAn(\L'(s),\X)$, it follows from the induction
hypothesis that $\psi\in\LQKXAn(\L(s),\X)$ and $(M,s)\not\sat \psi'$ iff
$\tau(\psi)\in\LQKXAn(\L'(s),\X)$ and $(M',s)\not\sat \tau(\psi')$. The
latter is true iff $(M',s)\sat \tau(\psi)$.
\item Suppose that $\psi=\psi_1\land\psi_2$. Then, $(M,s)\sat \psi$ iff
$(M,s)\sat \psi_1$ and $(M,s)\sat \psi_2$. It follows from the induction
hypothesis that $(M,s)\sat \psi_1$ and $(M,s)\sat \psi_2$ iff
$(M',s)\sat \tau(\psi_1)$ and $(M',s)\sat \tau(\psi_2)$. The latter is
true iff $(M',s)\sat \tau(\psi)$.
\item Suppose that $\psi=K_i\psi'$. Then, $(M,s)\sat \psi$ iff
$\psi\in\LQKXAn(\L(s),\X)$ and $(M,t)\sat \psi'$ for
every $t$ such that $(s,t\in \K_i)$.
Since by definition, $\psi\in\LQKXAn(\L(s),\X)$ iff
$\tau(\psi)\in\LQKXAn(\L'(s),\X)$, it follows from the induction
hypothesis that $\psi\in\LQKXAn(\L(s),\X)$ and $(M,t)\sat \psi'$ for
every
$t$ such that $(s,t)\in\K_i$ iff $\tau(\psi)\in\LQKXAn(\L'(s),\X)$ and
$(M',t)\sat \tau(\psi')$ for every $t$ such that $(s,t)\in\K_i$.
The latter is true iff $(M',s)\sat \tau(\psi)$.
\item Suppose that $\psi=A_i\psi'$. Then, $(M,s)\sat \psi$ iff $\psi'\in
\A_i(s)$. By definition, the latter is true iff $\tau(\psi')\in\A'_i(s)$
which is equivalent to $(M',s)\sat A_i(\tau(\psi'))$. The latter is true
iff $(M',s)\sat \tau(\psi)$.
\item Suppose that $\psi=X_i\psi'$. Then, $(M,s)\sat \psi$ iff
$(M,s)\sat K_i\psi'$ and $(M,s)\sat A_i\psi'$. It follows from the
previous results that $(M,s)\sat K_i\psi'$ and $(M,s)\sat A_i\psi'$ iff
$(M',s)\sat \tau(K_i\psi')$ and $(M',s)\sat \tau(A_i\psi')$. The latter
is true iff $(M',s)\sat \tau(\psi)$.
\item Finally, suppose that $\psi=\forall x\psi'$. Then,
}
In this case, we have that
$(M,s)\sat \psi$ iff $(M,s)\sat \psi'[x/\beta]$ for
all $\beta\in\LKAn(\L(s))$. By the induction
hypothesis, $(M,s)\sat \psi'[x/\beta]$ for all
$\beta\in\LKAn(\L(s))$ iff $(M^\tau,s)\sat \tau(\psi'[x/\beta])$
for all $\beta\in\LKAn(\L(s))$. Since
$\tau(\psi'[x/\beta])=\tau(\psi')[x/\tau(\beta)]$ and,
by construction of $\L^\tau$,
for all $\gamma\in\LKAn(\L^\tau(s))$ there exists $\beta\in\LKAn(\L(s))$
such that $\gamma=\tau(\beta)$, it follows that
$(M^\tau,s)\sat \tau(\psi'[x/\beta])$
for all $\beta\in\LKAn(\L(s))$ iff
$(M^\tau,s)\sat \tau(\psi')[x/\gamma])$ for all
$\gamma\in\LKAn(\L^\tau(s))$. The latter statement is true iff
$(M^\tau,s)\sat
\tau(\psi)$.
\commentout{
To finish the proof of the claim, suppose now that $\psi$ is an arbitrary formula whose set of free variables is $\{x_1,\ldots,x_k\}$. Then, $(M,s,\V)\sat \psi$ iff $(M,s,\V)\sat \psi[x_1/\V(s,x_1),\ldots,x_k/\V(s,x_k)]$. Since $\psi[x_1/\V(s,x_1),\ldots,x_k/\V(s,x_k)]$ is a sentence, the above argument implies that $(M,s,\V)\sat \psi[x_1/\V(s,x_1),\ldots,x_k/\V(s,x_k)]$ iff $(M',s,\V')\sat \tau(\psi[x_1/\V(s,x_1),\ldots,x_k/\V(s,x_k)])$. Since
\begin{eqnarray}
& & \tau(\psi[x_1/\V(s,x_1),\ldots,x_k/\V(s,x_k)])\nonumber \\
& & =\tau(\psi)[x_1/\tau(\V(s,x_1)),\ldots,x_k/\tau(\V(s,x_k))]\nonumber \\
& & =\tau(\psi)[x_1/\V'(s,x_1),\ldots,x_k/\V'(s,x_k)],\nonumber
\end{eqnarray}
the latter is true iff $(M',s\V')\sat \tau(\psi)$, as desired.
}

To complete the proof of the lemma, suppose that $\phi$ is a
sentence
that does not mention $q$ and that
$(M,s)\sat \phi$.
Let $\tau$ be a 1-1 function such that $\tau(p)=p$ for every $p$ that
occurs in $\phi$
and such that there exists no $r\in \Phi$ such that $\tau(r)=q$.
(Here we are using the fact that $\Phi$ is an infinite set.)
Note
that $\phi=\tau(\phi)$. Thus, the claim implies that
$(M',s)\sat \phi$
and by construction $q\notin\L'(s)$ for every $s\in S$.
\eprf

\emph{Substitution} is a standard property of most propositional logics.
It says that if $\phi$ is valid, then so is $\phi[q/\psi]$.
Substitution in full generality is not valid in our framework, because
of the semantics of quantification.  For example, although $\forall x
\neg A_i x \rimp \neg A_i q$ is valid, $\forall x \neg A_i x \rimp \neg A_i (\forall x A_i x)$
is not.   As we now show, if we restrict to quantifier-free
substitutions, we preserve validity.  But
this result depends on the fact
that $\Phi$ is infinite.  For example, if $\Phi = \{p,q\}$, then
$\phi = A_ip \land A_i q \rimp \forall x A_i x$ is valid, but $\phi[q/p]
= A_ip \land A_i p \rimp \forall x A_i x$ is not valid.  We first prove
that a slightly weaker version of Substitution holds (in which $q$
cannot appear in $\psi$), and then prove Substitution.
\pro {\bf (Weak Substitution)}
\label{lem:wsub}
If $\phi$ is a sentence valid in $\N_n(\Phi,\X)$, $q$ is a primitive proposition, and $\psi$ is an arbitrary
quantifier-free sentence that does not mention $q$, then $\phi[q/\psi]$ is valid in $\N_n(\Phi,\X)$.
\epro

\prf
Suppose, by way of contradiction, that $\phi[q/\psi]$ is not valid. Then
$\neg\phi[q/\psi]$ is satisfiable. By
Lemma~\ref{lem:satiswithoutq}, there exists an extended
awareness structure $M=(S,\L(s),\pi,\K_1,\ldots,\K_n,\A_1,\ldots,\A_n)$
and a world $s^*\in S$
such that
$(M,s^*)\sat \neg \phi[q/\psi]$
and $q\notin \L(s)$ for every $s\in S$.
Let $M'$ extends $M$ by defining $q$ as $\psi$; more precisely,
$M'=(S,\L',\pi',\K_1,\ldots,\K_n,\A'_1,\ldots,\A'_n)$, where
\begin{itemize}
\item $\L'(s)=\L(s)\union\{q\}$ if $\psi\in\LKAn(\L(s))$, and
$\L'(s)=\L(s)$ otherwise;
\item $\pi'(s,p)=\pi(s,p)$ for every $p\in\L(s)$ and if $q\in \L'(s)$, then
$\pi'(s,q)={\bf true}$
iff $(M,s)\sat\psi$;
\item $\A'_i(s)=\A_i(s)$ if $\psi\notin\A_i(s)$, and $\A'_i(s)$
is the smallest set generated by primitive propositions that includes
$\A_i(s)\union \{q\}$
otherwise.
\end{itemize}
Intuitively, we are just extending $M$ by defining $q$ so that it agrees
with $\psi$ everywhere.
We claim that for every
sentence
$\sigma$, if $\psi\in\LKAn(\L(s))$, then
the following are equivalent:
\begin{itemize}
\item[(a)]
$(M',s)\sat \sigma$
\item[(b)]
$(M',s)\sat \sigma[q/\psi]$
\item[(c)] $(M,s)\sat \sigma[q/\psi].$
\end{itemize}
We first observe that if $\sigma'$ is a quantifier-free sentence that
does not mention $q$, then for all worlds $s \in S$, we have that $(M,s) \sat
\sigma$ iff $(M',s) \sat \sigma'$.  (The formal proof is by a
straightforward induction on $\sigma'$.

We now prove the claim by induction in the structure of $\sigma$.  For
the base case, note that if $\sigma$ is the primitive proposition $q$,
then the equivalence between (b) and (c) follows from the observation
above.  All cases are straightforward except the case where $\sigma$ has
the form $\forall x \sigma'$.
\commentout{
\begin{itemize}
\item If $\sigma$ is a primitive proposition different from $q$, then $\sigma[q/\psi]=\sigma$ and the result follows from the definition of $\pi'$.
\item If $\sigma=q$, then by definition of $\pi'$ it follows that
$(M',s)\sat \sigma$ iff $(M,s)\sat \sigma[q/\psi]$. It remains to prove
that $(M',s)\sat \sigma[q/\psi]$ iff $(M,s)\sat \sigma[q/\psi]$. Since
$\sigma[q/\psi]=\psi$, we need to show that $(M',s)\sat \psi$ iff
$(M,s)\sat \psi$ for every $\psi$ that is a quantifier-free sentence
that does not mention $q$. We show this by induction on the size of
$\psi$.
    \begin{itemize}
    \item If $\psi$ is a primitive proposition, then the result follows from the definition of $\pi'$.
    \item Suppose that $\psi=\neg\psi'$. Then, $(M',s)\sat \neg\psi'$ iff $\psi\in\LKAn(\L'(s))$ and $(M',s)\not\sat\psi'$. Since $\psi$ does not mention $q$, it follows that $\psi\in\LKAn(\L'(s))$ iff $\psi\in\LKAn(\L(s))$. By the induction hypothesis, $(M',s)\not\sat\psi'$ iff $(M,s)\not\sat\psi'$. Thus, $\psi\in\LKAn(\L'(s))$ and $(M',s)\not\sat\psi'$ iff $\psi\in\LKAn(\L(s))$ and $(M,s)\not\sat\psi'$. The latter is true iff $(M,s)\sat \neg\psi'$.
    \item Suppose that $\psi=\psi_1\land\psi_2$. Then, $(M',s)\sat \psi$ iff $(M',s)\sat \psi_1$ and $(M',s)\sat \psi_2$. By the induction hypothesis, $(M',s)\sat \psi_1$ and $(M',s)\sat \psi_2$ iff $(M,s)\sat \psi_1$ and $(M,s)\sat \psi_2$. The latter is true iff $(M,s)\sat \psi$.
    \item Suppose that $\psi=K_i\psi'$. Then, $(M',s)\sat \psi$ iff $\psi\in\LKAn(\L'(s))$ and $(M',t)\sat\psi'$ for every
$t$ such that $(s,t)\in \K_i$.
Since $\psi$ does not mention $q$, it follows that $\psi\in\LKAn(\L'(s))$ iff $\psi\in\LKAn(\L(s))$. By the
induction hypothesis, $(M',t)\sat\psi'$ for every $t$ such that $(s,t)\in \K_i$ iff $(M,t)\sat\psi'$ for every $t$ such that $(s,t)\in \K_i$.
Thus, $\psi\in\LKAn(\L'(s))$ and $(M',t)\sat\psi'$ for every
$t$ such that $(s,t)\in \K_i$ iff $\psi\in\LKAn(\L(s))$ and $(M,t)\sat\psi'$ for every $t$ such that $(s,t)\in \K_i$. The latter is true iff
$(M,s)\sat \psi$.
    \item Suppose that $\psi=A_i\psi'$. Then, $(M',s)\sat \psi$ iff $\psi\in\LKAn(\L'(s))$ and $\psi'\in\A'_i(s)$. Since $\psi$ does not mention $q$, it follows that $\psi\in\LKAn(\L'(s))$ iff $\psi\in\LKAn(\L(s))$. Moreover, since $\psi$ does not mention $q$, by definition of $\A'_i$, it follows that $\psi'\in\A'_i(s)$ iff $\psi'\in\A_i(s)$. Thus, $\psi\in\LKAn(\L'(s))$ and $\psi'\in\A'_i(s)$ iff $\psi\in\LKAn(\L(s))$ and $\psi'\in\A_i(s)$. The latter is true iff $(M,s)\sat \psi$.
    \item Finally, suppose that $\psi=X_i\psi'$. Then, $(M',s)\sat \psi$ iff $(M',s)\sat K_i\psi'$ and $(M',s)\sat A_i\psi'$. By the previous cases, it follows that $(M',s)\sat K_i\psi'$ and $(M',s)\sat A_i\psi'$ iff $(M,s)\sat K_i\psi'$ and $(M,s)\sat A_i\psi'$. The latter is true iff $(M,s)\sat \psi$.
    \end{itemize}
\item Suppose that $\sigma=\neg\sigma'$. Then, $(M',s)\sat \neg\sigma'$ iff $\sigma \in\LQKXAn(\L'(s),\X)$ and $(M',s)\not\sat\sigma'$. Since $\psi\in \LKAn(\L(s))$ does not mention $q$, it follows that $\sigma\in\LQKXAn(\L'(s),\X)$ iff $\sigma[q/\psi]\in\LQKXAn(\L'(s),\X)$ iff $\sigma[q/\psi]\in\LQKXAn(\L(s),\X)$. By the induction hypothesis, $(M',s)\not\sat\sigma'$ iff $(M',s)\not\sat\sigma'[q/\psi]$ iff $(M,s)\not\sat\sigma'[q/\psi]$. Thus, $\sigma \in\LQKXAn(\L'(s),\X)$ and $(M',s)\not\sat\sigma'$ iff $\sigma[q/\psi]\in\LQKXAn(\L'(s),\X)$ and $(M',s)\not\sat\sigma'[q/\psi]$ iff $\sigma[q/\psi]\in\LQKXAn(\L(s),\X)$ and $(M,s)\not\sat\sigma'[q/\psi]$, as desired.
\item Suppose that $\sigma=\sigma_1\land\sigma_2$. Then, $(M',s)\sat \sigma$ iff $(M',s)\sat \sigma_1$ and $(M',s)\sat \sigma_2$. By the induction hypothesis, $(M',s)\sat \sigma_1$ and $(M',s)\sat \sigma_2$ iff $(M',s)\sat \sigma_1[q/\psi]$ and $(M',s)\sat \sigma_2[q/\psi]$ iff $(M,s)\sat \sigma_1[q/\psi]$ and $(M,s)\sat \sigma_2[q/\psi]$, as desired.
\item Suppose that $\sigma=K_i\sigma'$. Then, $(M,s)\sat \psi$ iff
$\sigma \in\LQKXAn(\L'(s),\X)$ and $(M',t)\sat \sigma'$ for every
    $t$ such that $(s,t)\in \K_i$.
    Since $\psi\in \LKAn(\L(s))$ does not mention $q$, it follows that $\sigma\in\LQKXAn(\L'(s),\X)$ iff $\sigma[q/\psi]\in\LQKXAn(\L'(s),\X)$ iff $\sigma[q/\psi]\in\LQKXAn(\L(s),\X)$. By the induction hypothesis, $(M',t)\sat \sigma'$ for every
    $t$ such that $(s,t)\in\K_i$ iff $(M',t)\sat \sigma'[q/\psi]$ for every $t$ such that $(s,t)\in\K_i$ iff $(M,t)\sat \sigma'[q/\psi]$ for every $t$ such that $(s,t)\in\K_i$. Thus, $\sigma \in\LQKXAn(\L'(s),\X)$ and $(M',t)\sat \sigma'$ for every $t$ such that $(s,t)\in\K_i$ iff $\sigma[q/\psi] \in\LQKXAn(\L'(s),\X)$ and $(M',t)\sat \sigma'[q/\psi]$ for every $t$ such that $(s,t)\in\K_i$ iff $\sigma[q/\psi] \in\LQKXAn(\L(s),\X)$ and $(M,t)\sat \sigma'[q/\psi]$ for every $t$ such that $(s,t)\in\K_i$, as desired.
\item Suppose that $\sigma=A_i\sigma'$. First, note that since $\psi\in \LKAn(\L(s))$ does not mention $q$, it follows that $\sigma\in\LQKXAn(\L'(s),\X)$ iff $\sigma[q/\psi]\in\LQKXAn(\L'(s),\X)$ iff $\sigma[q/\psi]\in\LQKXAn(\L(s),\X)$. Consider 4 possible cases:
    \begin{enumerate}
    \item Suppose that $\sigma'$ mentions $q$ and $\psi\in\A_i(s)$. Then, $q\in \A'_i(s)$ and $\sigma'\in\A'_i(s)$ iff $\sigma'[q/\psi]\in\A'_i(s)$ iff $\sigma'[q/\psi]\in\A_i(s)$.
    \item Suppose that $\sigma'$ mentions $q$ and $\psi\not\in\A_i(s)$. Then, $q\not\in \A'_i(s)$ and it follows that $\sigma'\not\in\A'_i(s)$, $\sigma'[q/\psi]\not\in\A'_i(s)$, and $\sigma'[q/\psi]\not\in\A_i(s)$.
    \item Suppose that $\sigma'$ does not mention $q$ and $\psi\in\A_i(s)$. Then, $q\in \A'_i(s)$, $\sigma'[q/\psi]=\sigma'$, and $\sigma'\in\A'_i(s)$ iff $\sigma'\in\A_i(s)$.
    \item Suppose that $\sigma'$ does not mention $q$ and $\psi\not\in\A_i(s)$. Then, $\A'_i(s)=\A_i(s)$, $\sigma'[q/\psi]=\sigma'$, and $\sigma'\in\A'_i(s)$ iff $\sigma'\in\A_i(s)$.
    \end{enumerate}
    Thus, in the 4 possible cases, it follows that $\sigma'\in\A'_i(s)$ iff $\sigma'[q/\psi]\in\A'_i(s)$ iff $\sigma'[q/\psi]\in\A_i(s)$. Thus, $(M',s)\sat \sigma$ iff $\sigma\in\LQKXAn(\L'(s),\X)$ and $\sigma'\in\A'_i(s)$ iff $\sigma[q/\psi]\in\LQKXAn(\L'(s),\X)$ and $\sigma'[q/\psi]\in\A'_i(s)$ iff $\sigma[q/\psi]\in\LQKXAn(\L(s),\X)$ and $\sigma'[q/\psi]\in\A_i(s)$, as desired.
\item Suppose that $\sigma=X_i\sigma'$. Then, $(M',s)\sat \sigma$ iff
$(M',s)\sat K_i\sigma'$ and $(M',s)\sat A_i\sigma'$. It follows from the
previous results that $(M',s)\sat K_i\sigma'$ and $(M',s)\sat
A_i\sigma'$ iff $(M',s)\sat K_i\sigma'[q/\psi]$ and $(M',s)\sat
A_i\sigma'[q/\psi]$ iff $(M,s)\sat K_i\sigma'[q/\psi]$ and $(M,s)\sat
A_i\sigma'[q/\psi]$, as desired.
\item Finally, suppose that $\sigma=\forall x\sigma'$. We divide the proof in two steps:
    \begin{enumerate}
}
To see that (a) implies (b),
suppose that $(M',s)\sat \forall x\sigma'$. Then $(M',s)\sat
\sigma'[x/\beta]$ for all
$\beta\in\LKAn(\L'(s))$. By the induction hypothesis,
$(M',s)\sat (\sigma'[x/\beta])[q/\psi]$.
Note that $\sigma'[x/\beta][q/\sigma] =
((\sigma'[q/\psi])[x/\beta])[q/\psi]$.  Thus, applying the induction
hypothesis again, it follows that
$(M',s) \sat (\sigma'[q/\psi])[x/\beta]$ for all $\beta \in \LKAn(\L'(s))$.
Therefore, $(M',s)\sat \forall x\sigma'[q/\psi]$.  This shows that (a)
implies (b).

To see that (b) implies (c),
suppose that $(M',s)\sat \forall x\sigma'[q/\psi]$. Thus,
$(M',s)\sat (\sigma'[q/\psi])[x/\beta]$ for all
$\beta\in\LKAn(\L'(s))$. Since
$\LKAn(\L(s))\subseteq \LKAn(\L'(s))$, by the induction hypothesis,
it follows that $(M,s)\sat (\sigma'[q/\psi])[x/\beta]$ for
all
$\beta\in\LKAn(\L(s))$. Thus, $(M,s)\sat
\forall x\sigma'[q/\psi]$.

Finally, to see that (c) implies (a),
suppose that $(M,s)\sat \forall x\sigma'[q/\psi]$.
We want to show that
$(M',s)\sat \forall x \sigma'$, or equivalently, that
$(M',s)\sat \sigma'[x/\beta]$ for all $\beta \in \LKAn(\L(s'))$.
Choose $\beta \in \LKAn(\L(s'))$.  So choose $\beta \in \LKAn(\L(s'))$.
By the induction hypothesis, $(M',s)\sat \sigma'[x/\beta]$ iff
$(M',s)\sat (\sigma'[x/\beta])[q/\psi]$ iff
$(M,s)\sat (\sigma'[x/\beta])[q/\psi]$.  Since
$(\sigma'[x/\beta])[q/\psi] = \sigma'[q/\sigma](x/\beta[q/\sigma])$,
and $(M,s) \sat \sigma'[q/\sigma](x/\beta[q/\sigma])$ since
$(M,s) \sat \forall x \sigma'[q/\sigma]$,  by assumption, the desired
result follows.
\commentout{
To finish the proof of the claim, suppose now that $\sigma$ is an arbitrary formula whose set of free variables is $\{x_1,\ldots,x_k\}$. Then, $(M',s,\V)\sat \sigma$ iff $(M',s,\V)\sat \sigma[x_1/\V(s,x_1),\ldots,x_k/\V(s,x_k)]$. Since $\sigma[x_1/\V(s,x_1),\ldots,x_k/\V(s,x_k)]$ is a sentence, the above argument implies that $(M',s,\V)\sat \psi[x_1/\V(s,x_1),\ldots,x_k/\V(s,x_k)]$ iff $(M',s,\V)\sat (\sigma[x_1/\V(s,x_1),\ldots,x_k/\V(s,x_k)])[q/\psi]$ iff $(M,s,\V)\sat (\sigma[x_1/\V(s,x_1),\ldots,x_k/\V(s,x_k)])[q/\psi]$. Finally, the result follows from the fact that $$(M',s,\V)\sat (\sigma[x_1/\V(s,x_1),\ldots,x_k/\V(s,x_k)])[q/\psi]\mbox{ iff }(M',s,\V)\sat \sigma[q/\psi], \mbox{ and }$$ $$(M,s,\V)\sat (\sigma[x_1/\V(s,x_1),\ldots,x_k/\V(s,x_k)])[q/\psi]\mbox{ iff }(M,s,\V)\sat \sigma[q/\psi].$$
}

Since, by assumption, $(M,s^*)\sat
\neg\phi[q/\psi]$, it follows from the claim above that $(M',s^*)\sat
\neg\phi$, a contradiction.
\eprf

\cor {\bf (Substitution)}
\label{cor:subst}
If $\phi$ is a sentence valid in $\N_n(\Phi,\X)$, $q$ is a primitive proposition, and $\psi$ is an arbitrary quantifier-free sentence, then $\phi[q/\psi]$ is valid in $\N_n(\Phi,\X)$.
\ecor

\prf
Choose a primitive
proposition $r$ that does not appear in $\psi$ or $\phi$.  By Weak
Substitution (Proposition~\ref{lem:wsub}), $\phi' = \phi[q/r]$ is valid.
Applying Weak Substitution
again, $\phi'[r/\psi] = \phi[q/\psi]$ is valid.
\eprf

We are finally ready to prove the soundness of MP.
\cor\label{MPsound} If $\phi \rimp \psi$ and $\phi$ are both valid in an
awareness structure $M$, then so is $\phi$.
\ecor

\prf Suppose, by way of contradiction, then $\phi \rimp \psi$ and $\phi$
are valid in $M$, and,
for some world $s$ in $M$,
we have that $(M,s) \sat \neg \phi$. It must be the
case that $\psi\notin \LQKXAn(\L(s),\X)$,
while $\phi\in \LQKXAn(\L(s),\X)$.  Let $q_1, \ldots, q_k$ be the
primitive propositions that are mentioned in $\psi$ but are not in
$\L(s)$.  Note that none of $q_1, \ldots, q_k$ can appear in
$\phi$. Since, by assumption,
$\L(s)$ is non-empty, let $p\in \L(s)$,
and let $\psi' =
\psi[q_1/p, \ldots, q_k/p]$. By Weak Substitution,
$\psi'$ and $\psi' \rimp \phi$ are valid.  But $\psi'$ and $\phi$ are in
$\LQKXAn(\L(s),\X)$. Thus, we must have
$(M,s) \sat \psi'$ and $(M,s) \sat \psi' \rimp \phi$,
so $(M,s) \sat \phi$, a contradiction.
\eprf

The following two results prove the soundness of Gen$_\forall$ and
Barcan$^*_X$.
\pro
{\bf (Gen$_\forall$)}
\label{thm:Genforall}
If $\phi$ is a valid sentence in
$\N_n(\Phi,\X)$ and $q$ is an arbitrary primitive proposition,
then $\forall x\phi[q/x]$ is valid in $\N_n(\Phi,\X)$.
\epro

\prf
Suppose not. Then there exists an extended awareness structure in $M\in
\N_n(\Phi,\X)$ and a world $s$ such
that
$(M,s)\sat \neg \forall x \phi[q/x]$.
Thus, there exists a formula $\psi\in \LKAn(\L(s))$
such that
$(M,s)\sat \neg (\phi[q/x])[x/\psi]$.
Thus, $\phi[q/\psi]$ is not valid. By Substitution, it follows that $\phi$ is
not valid either, a contradiction.
\eprf

\pro
{\bf (Barcan$^*_X$)}
\label{thm:WB} $(A_i  (\forall x \phi) \land \forall x (A_i x \rimp X_i \phi)) \rimp X_i (\forall x A_i x \rimp \forall x \phi)$ is valid
in $\N_n(\Phi,\X)$.
\epro

\prf
Suppose that $(M,s) \sat  (A_i  (\forall x \phi)\land \forall x (A_i
x \rimp X_i \phi))$. Since awareness is generated by primitive
propositions, $(M,s) \sat A_i (\forall x A_i x \rimp \forall x
\phi)$. Suppose, by way of contradiction, that $ (M,s) \sat \neg X_i
(\forall x A_i x \rimp \forall x \phi)$. Then there must exist some
world $t$ such that $(s,t)\in\K_i$ and $ (M,t) \sat \neg (\forall x
A_i x \rimp \forall x \phi)$. Thus, $(M,t) \sat \forall x A_i x$ and
$ (M,t) \sat \neg \forall x \phi$.
Since $ (M,t) \sat \neg \forall x \phi$, it follows that there exists
$\psi\in\LXAn(\L(t))$ such that $(M,t) \sat \neg \phi[x/\psi]$. Since
$(M,t) \sat \forall x A_i x$, we must have $(M,t) \sat A_i \psi$.
Since $\A_i(s) = \A_i(t)$, we also have $(M,s) \sat A_i \psi$.
Since $(M,s) \sat \forall x (A_i x \rimp X_i \phi)$, it follows that
$(M,s) \sat X_i \phi[x/\psi]$.  Thus, $(M,t) \sat \phi[x/\psi]$, a
contradiction.
\eprf

With these results in hand, we can now prove
Theorem~\ref{thm:compwithoutK}.  We repeat the theorem here for the
convenience of the reader.

\othm{thm:compwithoutK}
If $\C_X$ is a (possibly empty) subset of
$\{\rm{T}_X, 4_X, 5_X\}$ and $C$ is the corresponding subset of $\{r,t,
e\}$, then $\axXn \union \C_X$ is a sound and
complete axiomatization of the language $\LQXAn(\Phi,\X)$ with
respect to $\N_n^{C}(\Phi,\X)$.
\eothm

\prf
Corollary~\ref{MPsound} and Propositions~\ref{thm:Genforall} and
\ref{thm:WB} show the soundness of MP, Gen$_\forall$, and Barcan$^*_X$,
respectively.
The proof of soundness for the other axioms and rules is
standard and left to the reader.
The soundness of $\axXn \union \C_X$
follows easily.

We now consider completeness.
As we said in the main text, the proof is quite similar  in spirit to
that of Theorem~\ref{thm:awofunaaxiomswithoutK} given in HR.  We focus
here on the differences.
We give the remainder of the proof only for the case $\C_X=\emptyset$;
the other cases follow using standard techniques (see, for example,
\cite{FHMV,HC96}).

\commentout{
\lem \label{lem0} If ${\mathbb X}_n^{\forall}\union \C \vdash \phi$
and $x$ is substitutable for $q$ in $\phi$, then ${\mathbb X}_n^{\forall}\union \C \vdash \forall x \phi[q/x]$. \elem

\prf
We first show by induction on the length of the proof of $\phi$ that
if $z$ is a variable that does not appear in any formula in the
proof of $\phi$, then ${\mathbb X}_n^{\forall}\union \C \vdash
\phi[q/z]$. If there is a proof of $\phi$ of length one, then
$\phi$ is an instance of an axiom.
It is easy to see that $\phi[q/z]$ is an instance of the same axiom.
(We remark that it is important in the case of axioms N$_\forall$
and $1_{\forall}$ that $z$ does not occur in $\phi$.)
Suppose that the lemma holds for all $\phi'$ that have a proof of
length no greater than $k$, and suppose that $\phi$ has a proof of
length $k+1$ where $z$ does not occur in any formula of the proof.
If the last step of the proof of $\phi$ is an axiom, then $\phi$ is
an instance of an axiom, and we have already dealt with this case.
Otherwise,
the last step in the proof of $\phi$ is an application of either
MP, Gen$_X$, or Gen$_\forall$. We consider these in turn.

If MP is applied at the last step, then there exists some $\phi'$,
such that $\phi'$ and $\phi'\rimp \phi$ were
previously proved
and, by assumption, $z$ does not occur in any formula of their
proof. By the induction hypothesis, both $\phi'[q/z]$ and $(\phi'
\rimp \phi)[q/z] = \phi'[q/z] \rimp \phi[q/z]$ are provable. The
result now follows by an application of MP.

The argument for Gen$_X$ and Gen$_\forall$ is essentially identical,
so we consider them together. Suppose that Gen$_X$ (resp.,
Gen$_{\forall}$) is applied at the last step.
Then $\phi$ has the form $A_i\phi'\rimp X_i\phi'$ (resp., $\forall y\phi'$) and
there is a proof of length at most $k$ for $\phi'$ where $z$ does
not occur in any formula in the proof.
Thus, by the induction hypothesis, $\phi'[q/z]$ is provable.  By
applying Gen$_X$ (resp., Gen$_\forall$), it immediately follows that
$\phi[q/z]$ is provable.

This completes the proof that $\phi[q/z]$ is provable
if $z$ does not appear in any formula in the proof.  Since proofs are
finite and there are countably many variables, there must be some
variable $z$ that does not appear in any formula in the proof.  Thus,
for this choice of $z$, $\phi[q/z]$ is provable.
By applying
Gen$_\forall$, it follows that $\forall z \phi[q/z]$ is provable.
Since $x$ is substitutable for $q$ in $\phi$, $x$ must be
substitutable for $z$ in $\phi[q/z]$.  Thus, by applying the axiom
$1_\forall$ and MP, we can prove $\phi[q/x]$. The fact that $\forall
x \phi[q/x]$ is provable now follows from Gen$_\forall$. \eprf
}

As usual, the idea of the completeness proof is to construct a
canonical model $M^c$ where the worlds are maximal consistent
sets of sentences. It is then shown that if $s_V$ is the world
corresponding to the maximal consistent set $V$, then $(M^c,s_V)\sat
\varphi$ iff $\phi\in V$.  As observed in HR, this will not quite work
in the presence of quantification, since there may be a maximal
consistent set $V$ of sentences such that $\neg\forall x\phi\in V$,
but $\phi[x/\psi]$ for all $\psi\in \LKAn(\Phi)$. That is, there is
no witness to the falsity
of $\forall x \phi$ in $V$. This problem was dealt with in HR by
restricting to maximal consistent sets $V$ that are
\emph{acceptable} in the sense that if $\neg \forall x \phi \in V$,
then $\neg \phi[x/q] \in V$ for infinitely many primitive propositions
$q \in \Phi$.  (Note that this notion of acceptability also requires
$\Phi$ to be infinite.)   Because here we have possibly different languages
associated different worlds, we need to consider acceptability and
maximality with respect to a language.

\dfn
A set $\Gamma$ is {\em acceptable with respect to $L \subseteq \Phi$}
if $\phi\in\LQXAn(L,\X)$ and $\Gamma\vdash \phi[x/q]$ for all
but finitely many primitive propositions $q \in L$, then $\Gamma\vdash
\forall x \phi$.
\edfn

\dfn
If AX is an axiom system,
a set $\Gamma$ is {\em maximal AX-consistent set of sentences with
respect to $L \subseteq \Phi$}
if $\Gamma$ is a set of sentences contained in $\LQXAn(L,\X)$ and, for
all sentences $\phi\in \LQXAn(L,\X)$,
if $\Gamma\union\{\phi\}$
is $AX$-consistent, then $\phi\in\Gamma$.
\edfn

The following four lemmas are essentially Lemmas A.4, A.5, A.6, and A.7 in
HR.  Since the proofs are essentially identical, we do not repeat them
here.

\lem
\label{Claim4K}
If $\Gamma$ is a finite
set of sentences,
then $\Gamma$ is
acceptable with respect to every subset $L \subseteq \Phi$ that contains
infinitely
many primitive propositions.
\elem

\commentout{
\prf
Let $\Gamma=\{\beta_1,\ldots,\beta_k\}$ and $\beta=\beta_1\land\cdots\land\beta_k$. We need to show that if
$\phi\in\LQXAn(L,\X)$ and
$${\mathbb X}_n^{\forall}\vdash \beta\Rightarrow \phi[x/q],$$
for all but finitely many primitive propositions $q\in L$, then
$${\mathbb X}_n^{\forall}\vdash \beta\Rightarrow \forall x\phi.$$
Let $q$ be a primitive proposition in $L$ not occurring in $\Gamma\cup\{\phi\}$ such that
$${\mathbb X}_n^{\forall}\vdash \beta\Rightarrow \phi[x/q].$$
(Since there are infinitely many $q$'s in $L$ such that ${\mathbb X}_n^{\forall}\vdash \beta\Rightarrow \phi[x/q]$ and $\Gamma$ is finite, it is always possible to pick one $q$ that does not occur in $\Gamma\cup\{\phi\}$.)

By Gen$_\forall$, we have
$${\mathbb X}_n^{\forall}\vdash \forall x(\beta\Rightarrow \phi).$$
Since $\beta$ is a sentence, applying $K_\forall$ and $N_\forall$, it easily follows that
$${\mathbb X}_n^{\forall}\vdash \beta\Rightarrow \forall x\phi.$$
\eprf
}

\lem
\label{Claim5K}
If $\Gamma$ is
acceptable with respect to $L$
and $\tau$ is a sentence
in $\LQXAn(L,\X)$,
then $\Gamma \union \{\tau\}$ is
acceptable with respect to $L$.
\elem

\commentout{
\prf
Let $\Gamma' = \Gamma \union \{\tau\}$.
Suppose that
$\phi\in \LQXAn(l,\X)$ and
$\Gamma' \vdash \phi[x/q]$ for all but finitely many primitive
propositions $q\in L$. Then
$\Gamma \vdash \tau \Rightarrow \phi[x/q]$ for all but finitely many
$q\in L$, so
$\Gamma \vdash \forall x (\tau \Rightarrow \phi)$, since $\Gamma$ is
acceptable
with respect to $L$ and $(\tau \Rightarrow \phi)\in \LQXAn(L,\X)$.
Since $\tau$ is a sentence, applying $K_\forall$ and $N_\forall$, it easily follows that
$${\mathbb X}_n^{\forall}\vdash (\forall x (\tau \Rightarrow \phi)) \Rightarrow (\tau\Rightarrow \forall x\phi).$$
Thus, we have that $\Gamma \vdash \tau\Rightarrow \forall x\phi$,
so $\Gamma' \vdash \forall x \phi$, as desired.
\eprf
}
\commentout{
\lem
\label{Claim2.5X}
If $\Gamma$ is an acceptable with respect to $L$ maximal ${\mathbb
X}_n^{\forall}$-consistent set of sentences, then either (1)
$A_i\psi\in\Gamma$ for every sentence $\psi\in \LQXAn(\Phi,\X)$ or (2)
there are infinitely many primitive propositions $q\in L$ such that
$\neg A_iq\in\Gamma$. \elem

\prf
Suppose (2) is not the case, i.e, suppose that there are only finitely
many primitive propositions $q\in L$ such that $\neg
A_iq\in\Gamma$. Since $\Gamma$ is maximal ${\mathbb X}_n^{\forall}$-consistent set of sentences, it follows that
$\Gamma\vdash A_iq$ for all
but finitely many $q$'s in $L$. Since $\Gamma$ is acceptable, it follows
that $\Gamma\vdash\forall xA_ix$. Thus, by 1$_{\forall}$, it follows
that $\Gamma\vdash A_iq$ for every $q\in\Phi$. Thus, since awareness is
generated by primitive propositions, it follows that $A_i\psi\in\Gamma$
for every sentence $\psi\in \LQXAn(\Phi,\X)$. Thus, (1) holds.
\eprf
}

\commentout{
\lem
\label{Claim3X}
If $\Gamma$ is an acceptable maximal ${\mathbb X}_n^{\forall}\union \C_X$-consistent
set of sentences with respect to an infinite and co-infinite set $L
\subseteq \Phi$,
then there exists an infinite, co-infinite subset $L'$ of
$\Phi$ such that $\Gamma/X_i \union \{\neg A_i q : q \in L', A_iq \notin
\Gamma\}$ is
acceptable with respect to $L'$.
\elem

\prf
If $\Gamma/X_i \vdash \forall x  A_i x$, then define $L' = \{q: A_i q
\in \Gamma\}$; otherwise, define $L' = \{q: A_i q \in
\Gamma\}\union L''$, where $L''$ is an
infinite and co-infinite set of primitive propositions that do not occur
in $\Gamma$ (which exists, since, by assumption, $\Phi-L$
is infinite).
Note that $L'$ is infinite and co-infinite.  This is true
by construction if $\Gamma/X_i \not\vdash \forall x A_i x$.  If
$\Gamma/X_i \vdash \forall x  A_i x$, then it must be the case that $L'
= \{q: A_i q \in \Gamma\}$, for if it were finite, then we would have
that $\Gamma \vdash A_i q$ for only finitely many primitive propositions
in $L$.  Since $\Gamma$ is a maximal consistent set, it must be the case
that $\Gamma \vdash \neg A_i q $ for all but finitely many primitive
propositions $q \in L$.  Thus, since $\Gamma$ is acceptable, $\Gamma
\vdash \forall x \neg A_i x$.  But this contradicts the nonemptiness
axiom.  Thus, $L'$ must be infinite.  Since it is a subset of $L$, it is
clearly co-infinite, since $L$ is.
\commentout{
To see that $\Gamma/X_i$ is acceptable,
suppose that $\Gamma/X_i \vdash \phi[x/q]$ for
all but finitely many $q\in L'$.  There are two cases.  If $\Gamma/X_i \vdash \forall x A_i x$, then $\Gamma \vdash X_i \forall x A_i x$, and
$\Gamma \vdash X_i ( \phi[x/q])$ for all but finitely many $q$'s such
that $A_i q \in \Gamma$.  Thus, $\Gamma \vdash A_i q \rimp X_i
\phi[x/q]$ for all but finitely many $q$'s.    Since $\Gamma$ is
acceptable with respect to $L$, we have that $\Gamma \vdash
\forall x (A_i x \rimp X_i \phi)$.  Since $\Gamma/X_i \vdash \forall x
A_i x$, 1$\forall$ implies that $\Gamma/X_i \vdash A_i q$ for every $q$
that occurs in $\phi$. Since awareness is generated by primitive
propositions, it follows that $\Gamma/X_i \vdash A_i \forall
x\phi$. Thus, $\Gamma\vdash X_i A_i \forall x\phi$.
By A0$_X$ and AGPP, it easily follows that $\Gamma\vdash A_i \forall
x\phi$.
By Barcan$^*_X$, it
follows that $\Gamma \vdash X_i(\forall x A_i x \rimp  \forall x \phi)$.
Since $\Gamma \vdash X_i (\forall x A_i x)$, it follows that $\Gamma
\vdash X_i \forall x \phi$. Thus, $\Gamma/X_i \vdash \forall x \phi$.
In case (2), there must be some $q\in L'$ not mentioned in $\Gamma/X_i$
or $\phi$ such that $\Gamma/X_i \vdash \phi[x/q]$.
Since $\Gamma/X_i\vdash \phi[x/q]$, it follows that there exists a subset
$\{\beta_1\ldots,\beta_n\}\subseteq\Gamma/X_i$ such that
$${\mathbb X}_n^{\forall}\union \C_X\vdash \beta\Rightarrow \phi[x/q],$$
where $\beta=\beta_1\land\cdots\land\beta_n$.
Since $q$ does not occur in $\beta$ or $\phi$,
by Gen$_\forall$, we have
$${\mathbb X}_n^{\forall}\union \C_X\vdash \forall x(\beta\Rightarrow \phi).$$
Since $\beta$ is a sentence, applying $K_\forall$ and $N_\forall$, it easily follows that
$${\mathbb X}_n^{\forall}\union \C_X\vdash \beta\Rightarrow \forall x\phi,$$
which implies that $\Gamma/X_i\vdash \forall x\phi$, as desired. Thus,
the result also holds in case (2).
}

We need to consider two cases: (1) $\Gamma/X_i \vdash \forall x A_i x$
and (2) $\Gamma/X_i \not\vdash \forall x A_i x$.  In case (1) note that
$\{\neg A_i q : q \in L', A_iq \notin
\Gamma\}=\emptyset$. Thus, we need to show that $\Gamma/X_i$ is acceptable with respect to $L'$.
Suppose that $\Gamma/X_i \vdash \phi[x/q]$ for
all but finitely many $q\in L'$.  Since $\Gamma/X_i \vdash \forall x A_i x$, we have that $\Gamma \vdash X_i \forall x A_i x$, and
$\Gamma \vdash X_i ( \phi[x/q])$ for all but finitely many $q$'s such
that $A_i q \in \Gamma$.  Thus, $\Gamma \vdash A_i q \rimp X_i
\phi[x/q]$ for all but finitely many $q$'s.    Since $\Gamma$ is
acceptable with respect to $L$, we have that $\Gamma \vdash
\forall x (A_i x \rimp X_i \phi)$.  Since $\Gamma/X_i \vdash \forall x
A_i x$, 1$\forall$ implies that $\Gamma/X_i \vdash A_i q$ for every $q$
that occurs in $\phi$. Since awareness is generated by primitive
propositions, it follows that $\Gamma/X_i \vdash A_i \forall
x\phi$. Thus, $\Gamma\vdash X_i A_i \forall x\phi$.
By A0$_X$ and AGPP, it easily follows that $\Gamma\vdash A_i \forall
x\phi$.
By Barcan$^*_X$, it
follows that $\Gamma \vdash X_i(\forall x A_i x \rimp  \forall x \phi)$.
Since $\Gamma \vdash X_i (\forall x A_i x)$, it follows that $\Gamma
\vdash X_i \forall x \phi$. Thus, $\Gamma/X_i \vdash \forall x \phi$.

In case (2), suppose that
$$\Gamma/X_i \union \{\neg A_i q : q \in L', A_iq \notin
\Gamma\}\vdash \phi[x/p],$$
for all but finitely many $p\in L'$.
By definition of $L'$, there must be some $q^*\in L'$ not mentioned in $\Gamma/X_i$
or $\phi$ such that
$$\Gamma/X_i \union \{\neg A_i q : q \in L', A_iq \notin
\Gamma\}\vdash \phi[x/q^*].$$
Since $\Gamma/X_i \union \{\neg A_i q : q \in L', A_iq \notin
\Gamma\}\vdash \phi[x/q^*]$, it follows that there exist subsets
$\{\beta_1\ldots,\beta_n\}\subseteq\Gamma/X_i$ and $\{\beta'_1\ldots,\beta'_m\}\subseteq\{\neg A_i q : q \in L', A_iq \notin
\Gamma\}$ such that
$${\mathbb X}_n^{\forall}\union \C_X\vdash \beta\Rightarrow \phi[x/q^*],$$
where
$\beta=\beta_1\land\cdots\land\beta_n\land\beta'_1\land\cdots\land\beta'_m$.
There are two cases to consider. If there is no $\beta'_k=\neg A_i q^*$, then $q^*$ does not occur in $\beta$ or $\phi$. Thus,
by Gen$_\forall$, we have
$${\mathbb X}_n^{\forall}\union \C_X\vdash \forall x(\beta\Rightarrow \phi).$$
Since $\beta$ is a sentence, applying $K_\forall$ and $N_\forall$, it easily follows that
$${\mathbb X}_n^{\forall}\union \C_X\vdash \beta\Rightarrow \forall x\phi,$$
which implies that $\Gamma/X_i\union \{\neg A_i q : q \in L', A_iq \notin
\Gamma\}\vdash \forall x\phi$, as desired. Now consider the case, where
there exists some $\beta'_k=\neg A_iq^*$. Define $\beta'$ to be the
formula such that is identical to $\beta$ except from removing the
conjunct $\neg A_iq^*$. Note that $q^*$ does not occur in $\beta'$ or
$\phi$.
Thus, it follows that,
$${\mathbb X}_n^{\forall}\union \C_X\vdash \beta'\Rightarrow (\neg A_iq^*\rimp \phi[x/q^*]),$$
and
applying Gen$_\forall$, we have
$${\mathbb X}_n^{\forall}\union \C_X\vdash \forall x (\beta'\Rightarrow (\neg A_i x \rimp \phi)).$$
Since $\beta'$ is a sentence, applying $K_\forall$ and $N_\forall$, it easily follows that
$${\mathbb X}_n^{\forall}\union \C_X\vdash \beta'\Rightarrow \forall x(\neg A_i x \rimp \phi).$$
Applying $K_\forall$, it easily follows that
$${\mathbb X}_n^{\forall}\union \C_X\vdash \beta'\Rightarrow (\forall x \neg A_i x \rimp \forall x\phi).$$
By the nonempitness axiom, it follows that
$${\mathbb X}_n^{\forall}\union \C_X\vdash \neg(\forall x \neg A_i x).$$
\eprf
}

\lem
\label{Claim6X}
If $\Gamma\subseteq \LQXAn(L,\X)$ is
an acceptable $\axXn$-consistent
set of sentences with respect to $L$,
then
$\Gamma$ can be extended to a set of sentences that is acceptable and
maximal
$\axXn$-consistent with respect to $L$.
\elem

\commentout{
\prf
Let
$\psi_1,\psi_2,\ldots$ be an enumeration of
the set of sentences in $\LQXAn(L,\X)$
such that if $\psi_k$ is of the form
$\neg\forall x\phi$, then there must exist a $j<k$ such that $\psi_j$ is
of the form $\forall x\phi$.
We construct a sequence $\Delta_0, \Delta_1, \ldots$ of
acceptable ${\mathbb X}_n^{\forall}$-consistent
sets
with respect to
$L$ such that (1) $\Delta_0 = \Gamma$;
(2) $\Delta_k \subseteq \Delta_{k+1}$ for all $k \ge 0$;
(3) for $k \ge 1$, either $\psi_k \in \Delta_k$ or $\Delta_k \vdash \neg
\psi_k$; and
(4)  for all $k \ge 1$ and $0< j< k$, if $\psi_j$ has the form $\forall
x \phi$ and
$\Delta_{j-1} \vdash \neg \forall x \phi$, then there
exist at least $k-j$ distinct primitive propositions $q_{j,1}, \ldots,
q_{j,k-j}$ in $L$ such that $\{\neg \phi[x/q_{j,1}], \ldots, \neg
\phi[x/q_{j,k-j}]\} \subseteq \Delta_k$.

We proceed by induction.
Suppose that we have constructed
$\Delta_0, \ldots, \Delta_{k-1}$  satisfying properties (1)--(4).
To construct $\Delta_k$,  we first add to
$\Delta_{k-1}$ one witness $\neg\phi_j[x/q]$ for every formula $\psi_j$
of the form $\forall x\phi_j$ that is not ${\mathbb X}_n^{\forall}$-consistent with
$\Delta_{k-1}$, for $1 \le j \le k-1$.
Formally, we inductively construct a sequence $\Delta_{k-1,0},
\Delta_{k-1,1},
\ldots,\Delta_{k-1,k-1}$ of
sets that are ${\mathbb X}_n^{\forall}$-consistent and acceptable with
respect to $L$, where  we add the
``witness'' for $\psi_j$, if necessary, at the $j$th step of the
construction.
Let
$\Delta_{k-1,0} = \Delta_{k-1}$.
For $1 \le j \le k-1$, suppose that we have defined
sets $\Delta_{k-1,j-1}$ that are acceptable and consistent with respect
to $L$.
If it is not the case that $\psi_j$ has the form
$\forall x \phi$, $\Delta_{k-1} \vdash \neg \psi_j$, and there are only
finitely many primitive propositions $q\in L$ such $\Delta_{k-1} \vdash \neg
\phi[x/q]$, then $\Delta_{k-1,j} = \Delta_{k-1,j-1}$.  Otherwise, since
$\Delta_{k-1,j-1}$ is acceptable with respect to
$L$,
there must be infinitely many primitive propositions $q\in L$ such that
$\Delta_{k-1,j-1} \union \{\neg \phi[x/q]\}$ is ${\mathbb X}_n^{\forall}$-consistent.  Choose $q\in L$
such that $\neg \phi[x/q] \notin \Delta_{k-1,j-1}$, and let
$\Delta_{k-1,j} = \Delta_{k-1,j-1} \union \{\neg \phi[x/q]\}$.
By Lemma~\ref{Claim5K}, $\Delta_{k-1,j}$ is acceptable with respect to
$L$.
Let $\Delta'_{k-1}=\Delta_{k-1,k-1}$;
$\Delta'_{k-1}$ has the required witnesses.
If $\Delta'_{k-1}\cup\{\psi_k\}$ is ${\mathbb X}_n^{\forall}$-consistent, then
$\Delta_k=\Delta'_{k-1}\cup\{\psi_k\}$.
By Lemma~\ref{Claim5K}, $\Delta_k$ is
acceptable with respect to $L$ and ${\mathbb X}_n^{\forall}$-consistent.
If $\Delta'_{k-1}\cup\{\psi_k\}$
is not ${\mathbb X}_n^{\forall}$-consistent, then $\Delta_k=\Delta'_{k-1}$.
Clearly this construction satisfies properties (1)--(4).

Let $\Delta=\cup_k \Delta_k$. Since $\Gamma\subseteq \LQXAn(L,\X)$,
clearly, $\Delta$ is a set of sentences that is maximal ${\mathbb X}_n^{\forall}$-consistent
with respect to $L$ and includes
$\Gamma$. Thus, it remains to verify that it is
acceptable with respect to $L$.
Suppose that
$\phi\in \LQXAn(L,\X)$ and
$\Delta \vdash \phi[x/q]$ for all but finitely many
primitive propositions $q\in L$. Since $\Delta$ is
a set of sentences that is maximal ${\bf
X}_n^{\forall}$-consistent with respect to $L$, it
follows that $\phi[x/q]\in \Delta$
for all but finitely many $q$'s in $L$.
Suppose that the formula $\psi_k$ is $\forall x \phi$.  By construction,
either $\psi_k \in \Delta_k$ (and hence in $\Delta$), or $\neg \phi[x/q]
\in \Delta$ for infinitely many primitive propositions $q$ in $L$.  The latter cannot be the case, since $\Delta$ is ${\mathbb X}_n^{\forall}$-consistent
and $\phi[x/q] \in \Delta$ for all but finitely many primitive propositions $q\in L$.  Thus,
$\Delta \vdash \forall x \phi$, as desired.
\eprf
}
Let $\Gamma/X_i=\{\phi:X_i\phi\in\Gamma\}$.
\lem
\label{LemmaA3X}
If $\Gamma$ is a
a set of sentences that is
maximal $\axXn$-consistent with respect to $L$
containing $\neg
X_i\phi$ and $A_i\phi$, then $\Gamma/X_i\cup\{\neg\phi\}$ is
$\axXn$-consistent.
\elem

Lemma A.14 in HR shows that if $\Gamma$ is an acceptable maximal
consistent set that contains $A_i \phi$ and $\neg X_i \phi$, then
$\Gamma/X_i \union \{\neg \phi\}$ can be extended to an acceptable
maximal consistent set $\Delta$.  (Lemma A.8 proves a similar result for
the $K_i$ operator.)
The following lemma proves an analogous result,
but here we must work harder to take the language into account.
That is, we have to define the language $L'$ with respect to which
$\Delta$ is maximal and acceptable.
As usual, we say that $L$ is \emph{co-infinite} if $\Phi - L$ is
infinite.

\lem
\label{LemmaA4X}
If $\Gamma$ is
an acceptable maximal $\axXn$-consistent set of
sentences with respect to $L$, where
$L$ is infinite and co-infinite,
$\neg X_i\phi\in \Gamma$, and $A_i\phi\in\Gamma$, then there
exist an infinite and co-infinite set $L'\subseteq\Phi$ and a
set $\Delta$ of sentences
that is
acceptable,
maximal $\axXn$-consistent with respect to $L'$ and
contains
$\Gamma/X_i \union \{\neg\phi\}$.
Moreover, $A_i\psi \in \Delta$ iff $A_i\psi\in\Gamma$ for
all formulas $\psi$.
\elem

\prf
By Lemma~\ref{LemmaA3X}, $\Gamma/X_i\cup\{\neg\phi\}$ is
$\axXn$-consistent.
We define a subset $L'\subseteq\Phi$ and construct a set $\Delta$ of
sentences that is  acceptable and maximal
$\axXn$-consistent
with respect to $L'$ such that $\Delta$
contains $\Gamma/X_i\cup\{\neg\phi\}$ and $A_i\phi \in \Delta$ iff
$A_i\phi\in\Gamma$ for all formulas $\phi$.

We consider two cases: (1) $\Gamma/X_i \union \{\neg \phi \} \vdash \forall
xA_i x$; and (2) $\Gamma/X_i \union \{\neg \phi \} \not\vdash \forall
xA_i x$.

If $\Gamma/X_i \union \{\neg \phi \} \vdash \forall xA_i x$,
then define $L'=\{q:A_iq\in\Gamma\}$.
Note that since $\Gamma \vdash A_i\phi$, it follows that every primitive
proposition $q$ in $\phi$ must be in $L'$, as is every primitive
proposition in a formula in $\Gamma/X_i$.
$L'$ must be infinite, for if it were finite,
then we would have
that $\Gamma \vdash A_i q$ for only finitely many primitive propositions
in $L$.  Since $\Gamma$ is a maximal
$\axXn$-consistent set, it must be the case
that $\Gamma \vdash \neg A_i q $ for all but finitely many primitive
propositions $q \in L$.  Since $\Gamma$ is acceptable
with respect to $L$,
$\Gamma
\vdash \forall x \neg A_i x$.
Thus, axiom FA$_X^*$ implies that $\forall x \neg A_i x\in \Gamma/X_i$, which
is a contradiction, since by assumption
$\Gamma/X_i \union\{\neg \phi\}\vdash \forall x A_i x$.
Thus, $L'$ must be infinite.  Since $L'$ is a subset of $L$, it is
clearly co-infinite, since $L$ is.

We prove that $\Gamma/X_i \union \{\neg \phi \}$ is acceptable with
respect to $L'$
in this case.
Suppose that
$\psi\in\LQXAn(L',\X)$ and
\begin{equation}\label{eq1}
\mbox{$\Gamma/X_i \union \{\neg \phi \}\vdash \psi[x/q]$ for
all but finitely many $q\in L'$.}
\end{equation}
We want to show that
$\Gamma/X_i \union \{\neg \phi \}\vdash \forall x \psi$.
It follows from (\ref{eq1}) that $\Gamma/X_i \vdash \neg\phi\rimp
\psi[x/q]$ for all but finitely many $q\in L'$.
Since every primitive proposition in $\psi$ is in $L' = \{q: A_iq\in
\Gamma\}$,  and $A_i \phi \in \Gamma$, it easily follows that
$\Gamma \vdash X_i (\neg\phi\rimp \psi[x/q])$ for
all but finitely many $q \in L'$.  Since $L' = \{q: A_i q \in \Gamma\}$,
it follows that $\Gamma \vdash A_i q \rimp X_i
(\neg\phi\rimp \psi[x/q])$ for all but finitely many $q\in L$.    Since
$\Gamma$ is
acceptable with respect to $L$, we have that
\begin{equation}\label{eq2}\Gamma \vdash
\forall x (A_i x \rimp X_i(\neg\phi\rimp \psi)).
\end{equation}
Again using the fact that $\Gamma \vdash A_i q$ for all $q$ in $\psi$ and
$\Gamma \vdash A_i \phi$, from AGPP we have that
\begin{equation}\label{eq3}
\Gamma \vdash A_i \forall x (\neg \phi \rimp \psi).
\end{equation}
From Barcan$^*_X$, (\ref{eq2}), and (\ref{eq3}), it follows that
$\Gamma \vdash X_i(\forall x A_i x \rimp  \forall x (\neg\phi\rimp \psi))$.
Thus, $\Gamma/X_i \vdash \forall x A_i x \rimp  \forall x (\neg\phi\rimp
\psi)$.
Since $\Gamma/X_i \union \{\neg \phi \} \vdash \forall x A_i x$, it follows that $\Gamma/X_i \union \{\neg \phi \}
\vdash \forall x (\neg\phi\rimp \psi)$.
Since $\phi$ is a sentence, applying $K_\forall$ and $N_\forall$, it easily follows that
$\Gamma/X_i \union \{\neg \phi \}
\vdash \neg\phi\rimp \forall x\psi$.
Thus, $\Gamma/X_i \union \{\neg \phi \}\vdash \forall x \psi$, as desired.

Therefore, $\Gamma/X_i \union \{\neg \phi \}$ is a set of sentences that
is acceptable with respect to $L'$ and
$\axXn$-consistent. Thus, by Lemma~\ref{Claim6X},
there exists a set of sentences $\Delta$ containing $\Gamma/X_i \union
\{\neg \phi \}$ that is acceptable and maximal
$\axXn$-consistent with respect to
$L'$.
Finally, we prove that $A_i\psi\in\Gamma$ iff $A_i\psi\in\Delta$. First, suppose that $A_i\psi\in\Gamma$. Then,
XA
implies that $X_iA_i\psi\in\Gamma$. Thus, $A_i\psi\in\Gamma/X_i\subseteq
\Delta$. For the converse, suppose that $A_i\psi\in\Delta$.
Since $\psi \in \LQXAn(L',\X)$, it must be the case that $\Gamma \vdash
A_i q$ for every primitive proposition $q$ that appears in
$\psi$; thus $\Gamma \vdash A_i \psi$.

If $\Gamma/X_i\union \{\neg \phi \}\not\vdash
\forall xA_i x$, define $L'=\{q:A_iq\in\Gamma\}\union L''$, where
$L''$ is an infinite and
co-infinite set of primitive propositions not occurring in $\Gamma
\union \{\phi\}$ (which exists, since, by assumption, $\Phi-L$
is infinite). It can be easily seen that $L'$ is infinite and co-infinite.
Since $\Gamma/X_i\union \{\neg \phi \}$ is
$\axXn$-consistent,
$\Gamma/X_i\union \{\neg \phi \}\not\vdash \forall xA_i x$ implies that $\Gamma/X_i \union \{\neg\phi,\neg\forall xA_ix\}$
is $\axXn$-consistent.

To see
that $\Gamma/X_i \union \{\neg\phi\}$ is acceptable with
respect to $L'$, suppose that
$\psi\in\LQXAn(L',\X)$ and
$\Gamma/X_i \union \{\neg\phi\} \vdash \psi[x/q]$ for
all but finitely many $q\in L'$.  There must be some $q\in L'$ not
mentioned in $\Gamma/X_i$
or $\phi$ such that $\Gamma/X_i \union \{\neg\phi\} \vdash \psi[x/q]$.
Since $\Gamma/X_i\union \{\neg\phi\}\vdash \psi[x/q]$, it follows that there exists a subset
$\{\beta_1\ldots,\beta_n\}\subseteq\Gamma/X_i\union \{\neg\phi\}$ such that
$\axXn\vdash \beta \Rightarrow \psi[x/q]$,
where $\beta=\beta_1\land\cdots\land\beta_n$.
Since $q$ does not occur in $\beta$ or $\phi$,
by Gen$_\forall$, we have
$\axXn \vdash \forall x(\beta\Rightarrow \psi)$.
Since $\beta$ is a sentence, applying $K_\forall$ and $N_\forall$, it
easily follows that
$\axXn \vdash \beta\Rightarrow \forall x\psi$,
which implies that $\Gamma/X_i\union \{\neg\phi\}\vdash \forall x\psi$, as desired.
Finally, since $\Gamma/X_i \union \{\neg\phi\}$
is acceptable with respect to $L'$, Lemma~\ref{Claim5K} implies that
$\Gamma/X_i \union \{\neg\phi,\neg\forall xA_ix\}$ is acceptable with
respect to $L'$.

Let  $\psi_1,\psi_2,\ldots$ be an enumeration of
the set of sentences in $\LQXAn(L',\X)$
such that if $\psi_k$ is of the form
$\neg\forall x\phi$, then there must exist a $j<k$ such that $\psi_j$ is
of the form $\forall x\phi$ and if $\psi_k$ is a formula that contains a primitive proposition $q\in L''$, then there must exist a $j<k$ such that $\psi_j$ is
of the form $\neg A_i q$.
The construction continues exactly as in the proof of
Lemma~\ref{Claim6X}, where we take $\Delta_0=\Gamma/X_i \union \{\neg
\phi,\neg \forall x A_ix\}$. Note that by construction, if $\psi_j=\neg
A_i q$ for some $q\in L''$, then $q$ does not occur in
$\Delta_{j-1}'$. We claim that $\Delta_{j-1}'\union\{\neg A_i q\}$ is
$\axXn$-consistent. For suppose otherwise.  Then, as above, there
exists a subset $\{\beta_1,\ldots,\beta_n\}\subseteq \Delta_{j-1}'$ such
that
$\axXn\vdash \beta\Rightarrow \forall x A_i x$
Since $\{\beta_1,\ldots,\beta_n,\neg\forall x A_ix\}\subseteq
\Delta_{j-1}'$, it follows that $\Delta_{j-1}'$ is not
$\axXn$-consistent, a contradiction.

Therefore, $\Delta$ is a set of sentences that is acceptable and maximal
$\axXn$-consistent with respect to $L'$ and includes
$\Gamma/X_i\union {\neg \phi}\union\{\neg A_iq:q\in L''\}$. The proof
that $A_i\psi\in\Gamma$ implies $A_i\psi\in\Delta$
is identical to the first case. For the converse, suppose that
$A_i\psi\in\Delta$.
Then, by AGPP, $A_iq \in \Delta$ for all primitive propositions $q$ that
appear in $\psi$.  The construction of $\Delta$ guarantees that, for all
primitive propositions in $L'$, we have $A_iq \in \Delta$ iff $A_i q \in
\Gamma$.
Since
$\Gamma$ is maximal $\bf{X}_n^{\forall}$-consistent with respect to
$L$, AGPP implies that $A_i\psi\in\Gamma$.
\eprf

\lem
\label{LemmaA5X}
If $\varphi$ is a $\axXn$-consistent sentence, then
$\varphi$ is satisfiable
in $\N_n^{\agpp, \ka, \emptyset}(\Phi,\X)$.
\elem

\prf
As usual, we construct a canonical model where the worlds are maximal
consistent sets of formulas.  However, now the worlds must also
explicitly include the language.  For technical reasons, we also assume
that the language is infinite and coninfinite.

Let $M^{c}= (\Sigma,\L,{\cal K}_1,...,{\cal
K}_n,\A_1,\ldots,\A_n,\pi)$ be a canonical extended awareness structure
constructed as follows
\begin{itemize}
\item $\Sigma=\{(s_V,L):V$ is
a set of sentences that is acceptable and maximal
$\axXn$-consistent with respect to $L$, where
$L\subseteq\Phi$
is infinite and co-infinite\};

\item $\L((s_V,L))=L$;

\item $
\pi((s_V,L),p)= \left\{
\begin{array}{ll}
                                {\bf true} & \mbox{if $p \in V$}, \\
                                {\bf false} & \mbox{if $p\in (L-V)$}; \\
                                \end{array}
                            \right.
                            $

\item $\A_i((s_V,L))=\{\phi:A_i\phi\in V\}$;

\item $\K_i((s_V,L))= \{(s_W,L'):V/X_i\subseteq W\mbox{ and } A_i\phi\in W\mbox{ iff }A_i\phi\in
V\mbox{ for all formulas }\phi\}$.

\end{itemize}

We show that if
$\psi\in \LQXAn(L,\X)$ is a sentence, then
\begin{eqnarray}
\label{eq:consiffsatis1} (M^{c},(s_V,L))\sat\psi\mbox{\ \ iff\ \
}\psi\in V.
\end{eqnarray}
Note that this claim
suffices to prove Lemma~\ref{LemmaA5X} since, for all $L\subseteq\Phi$ that is infinite and co-infinite, if $\varphi\in \LQXAn(L,\X)$ is
a $\axXn$-consistent sentence, by Lemmas~\ref{Claim4K}
and \ref{Claim6X}, it
is contained in
a set of sentences that is acceptable and maximal
$\axXn$-consistent with respect to $L$.

We prove (\ref{eq:consiffsatis1}) by induction of the depth of
nesting of $\forall$, with a subinduction on the length of the
sentence.  The details are standard and left to the reader.
For the case of $X_i \phi$, we need Lemma~\ref{LemmaA4X}.

If $\phi$ is consistent, by Lemmas~\ref{Claim4K} and \ref{Claim6X}, then
$\phi$ there is a set $L \subseteq \Phi$ that is infinite and co-infinite
and contains $\Phi(\phi)$ and a set $V$ of sentences that is acceptable
and maximal $\axXn$-consistent with respect to $L$ such that $\phi \in
V$.  By the argument above, $(M,(s_V,L)) \sat \phi$, showing that $\phi$
is satisfiable, as desired. \eprf

\commentout{
The base case is if $\psi$ is a primitive proposition, in which case
(\ref{eq:consiffsatis1}) follows immediately from the definition of
$\pi$. For the inductive step, given $\psi\in \LQXAn(L,\X)$, suppose that
(\ref{eq:consiffsatis1}) holds for all formulas $\psi'$ such that
either the depth of nesting for $\forall$ in $\psi'$ is less than
that in $\psi$, or the depth of nesting is the same, and $\psi'$ is
shorter than $\psi$.  We proceed by cases on the form of $\psi$.
\begin{itemize}
\item If $\psi$ has the form $\neg\psi'$ or  $\psi_1\land\psi_2$, then the
result follows easily from the inductive hypothesis.
\item If $\psi$ has the
form $A_i \psi'$, then note that $\psi'$ is a sentence and
$(M^c,(s_V,L)) \sat A_i \psi'$ iff $\psi' \in \A_i((s_V,L))$ iff $A_i
\psi' \in V$.

\item If $\psi$ has the form $X_i \psi'$, then if $\psi\in V$, then $\psi'\in W$ for every $W$
such that $(s_W,L')\in \K_i((s_V,L))$. By the induction hypothesis,
$(M^c,(s_W,L'))\sat \psi'$ for every $(s_W,L')\in \K_i((s_V,L))$. By $A0_X$, we have
that $A_i\psi'\in V$. Thus, $\psi'\in\A_i((s_V,L))$ which implies that
$(M^c,(s_V,L))\sat \psi$. If $\psi\notin V$,
since $\psi\in\LQXAn(L,\X)$, it follows that
$\neg\psi\in V$.
If $A_i\psi'\notin V$,
then $\psi'\notin\A_i((s_V,L))$ which implies that $(M^c,(s_V,L))\not\sat
\psi$.  If $A_i\psi'\in V$ then, by Lemma~\ref{LemmaA4X}, there exist
$L'\subseteq\Phi$ infinite and co-infinite and $W$ a set of sentences
that is acceptable and maximal $\axXn$-consistent with
respect to $L'$ such that
$V/X_i\cup\{\neg\psi'\}\subseteq W$ and, moreover, $A_i\phi \in W$ iff $A_i\phi\in V$ for every formula $\phi$. Thus, $(s_W,L')\in \K_i((s_V,L))$ and, by the induction hypothesis,
$(M^c,(s_W,L'))\not\sat \psi'$. Thus, $(M^c,(s_V,L))\not\sat \psi$.

\item
Finally, suppose that $\psi=\forall x \psi'$. If $\psi\in V$ then,
by axiom $1_{\forall}$, $\psi'[x/\phi]\in V$ for all $\phi\in
\LXAn(L)$.
The depth of nesting of $\psi'[x/\phi]$ is less than
that of $\forall x \psi'$,  so by the induction hypothesis
$(M,(s_V,L))\sat\psi'[x/\phi]$ for all $\phi\in \LXAn(L)$. By
definition, $(M,(s_V,L))\sat\psi$, as desired. If $\psi\notin V$, then
since $\psi\in\LQXAn(L,\X)$, it follows that
$\neg\psi\in V$.
Since
$\psi\in\LQXAn(L,\X)$ and
$V$ is
a set of sentences that is acceptable and maximal
$\axXn$-consistent with respect to $L$ , there must exist a
exist a
primitive proposition
$q\in L$ such that $\neg\psi'[x/q]\in V$. By the induction
hypothesis, $(M^c,(s_V,L))\not\sat\psi'[x/q]$. Thus,
$(M^c,(s_V,L))\not\sat\psi$, as desired.
\end{itemize}
\eprf
}

To finish the completeness proof, suppose that $\phi$ is valid in
$\N^{\agpp, \ka, \emptyset}_{n}(\Phi,\X)$.
Since $\phi$ is a sentence, it follows that
$\neg\phi$ is a sentence and is not satisfiable in
$\N^{\agpp, \ka, \emptyset}_{n}(\Phi,\X)$. So, by Lemma~\ref{LemmaA5X},
$\neg\phi$ is not $\axXn$-consistent.
Thus, $\phi$ is provable in $\axXn$.
\eprf

\opro{pro:soundness}
\begin{itemize}
\item[(a)] XA$^*$ is valid in $\N_n^{t}(\Phi,\X)$.
\item[(b)] Barcan$^*$ is valid in $\N_n^{r,e}(\Phi,\X)$.
\item[(c)] FA$^*$ is valid in $\N_n^{r,e}(\Phi,\X)$.
\item[(d)] 5$^*$ is valid in $\N_n^{e}(\Phi,\X)$.
\end{itemize}
\eopro

\prf For part (a),
suppose that $(M,s)\sat A_i^*\phi$, where $M \in \N_n^t(\Phi,\X)$.
Thus, $(M,s) \sat K_i (\phi \lor \neg \phi)$.  Since the axiom 4 is
valid in structures in $\N_n^t(\Phi,\X)$, it follows that
$(M,s) \sat K_i K_i (\phi \lor \neg \phi)$, that is,
$(M,s) \sat K_i A_i^* \phi$.

For part (b),
suppose that $(M,s) \sat  A_i^*  (\forall x \phi)\land \forall x (A_i^*
 x \rimp K_i \phi)$, where $M \in \N_n^{r,e}(\Phi,\X)$.
It easily follows that
$(M,s) \sat A_i^* (\forall x A_i^* x \rimp \forall x
\phi)$. Suppose, by way of contradiction, that $ (M,s) \sat \neg K_i
(\forall x A_i^* x \rimp \forall x \phi)$. Then there must exist some
world $t$ such that $(s,t)\in\K_i$ and $ (M,t) \sat \neg (\forall x
A_i^* x \rimp \forall x \phi)$. Thus, $(M,t) \sat \forall x A_i^* x$ and
$ (M,t) \sat \neg \forall x \phi$. Since $ (M,t) \sat \neg \forall x
\phi$, it follows that there exists
$\psi\in\LKAn(\L(t))$ such that $(M,t) \sat \neg \phi[x/\psi]$.
Since $(M,t) \sat \forall x A_i^* x$, we must have $(M,t) \sat A_i^* \psi$.
Thus, for every world $u$ such that $(t,u)\in \K_i$, it follows that
$\psi\in \LKAn(\L(u))$.
Suppose that $(s,v) \in \K_i$.  Since $\K_i$ is Euclidean and $(s,t) \in
\K_i$, it follows that $(t,v) \in \K_i$ and, by the observation above,
that $\psi\in \LKAn(\L(v))$.
Since $\K_i$ is reflexive and Euclidean, it follows that $(t,s) \in
\K_i$, so the argument above also shows that $\psi\in \LKAn(\L(s))$.
Thus, $(M,s) \sat A_i^* \psi$.
Since $(M,s) \sat \forall x (A_i^* x \rimp K_i \phi)$, it follows that
$(M,s) \sat K_i \phi[x/\psi]$.  Thus, $(M,t) \sat \phi[x/\psi]$, a
contradiction.

Finally, for part (c),
suppose that $(M,s)\sat \forall x \neg A_i^* x$, where $M \in
\N_n^{r,e}(\Phi,\X)$. Thus, for every
primitive proposition $p\in \L(s)$, there exists some
$t_p$ such that
$(s,t_p)\in \K_i$ and $p\notin \L(t_p)$. Let $u$ be an arbitrary world such
that $(s,u)\in \K_i$. Let $\phi$ be an arbitrary quantifier-free
sentence in $\LQKXAn(\L(u),\X)$.
If $\Phi(\phi)\inter \L(s) \ne \emptyset$, suppose that $p \in \Phi(\phi)
\inter \L(s)$.  By assumption, $p \notin \L(t_p)$.  Since $\K_i$ is
Euclidean, $(u,t_p) \in \K_i$.  Thus, $(M,u) \sat \neg A_i^* \phi$.
If $\Phi(\phi)\inter \L(s) = \emptyset$, note that since $\K_i$ is
reflexive and Euclidean, the fact that $(s,s)$ and $(s,u)$ are in $\K_i$
implies that $(u,s) \in \K_i$.  Hence, we again have that $(M,u) \sat
\neg A_i^* \phi$.
The proof of part (d) is standard, and left to the reader.
\eprf

\othm{thm:compwithK}
\begin{itemize}
\item[(a)] $\axKn\union\{\rm{T},4,5^*\}$ is a sound and complete
axiomatization of
the sentences in $\LQKXAn(\Phi,\X)$ with respect to $\N_n^{r,t,e}(\Phi,\X)$.
\item[(b)] $\axKnr\union\{\rm{T},4,5^*\}$ is a sound and complete axiomatization of
the sentences in $\LQKn(\Phi,\X)$ with respect to $\N_n^{r,t,e}(\Phi,\X)$.
\item[(c)] $\axKnrr\union\{\rm{T},4,5^*\}$ is a sound and complete axiomatization of
$\LKn(\Phi)$ with respect to $\N_n^{r,t,e}(\Phi)$.
\end{itemize}
\eothm

\prf
The proof of part (a) is identical to the proof of
Theorem~\ref{thm:compwithoutK}, except that
$X_i$ and $A_i$ are replaced by $K_i$ and $A_i^*$, respectively, and in
Lemma~\ref{LemmaA5X}, another step is needed in the induction to deal with
$X_i$ that uses the extra axiom A0 in the standard way.

For part (b), note that since $X_i$ and $A_i$ are not part of the language the axioms of $\axKn$ that mention these operators are not needed in the induction of Lemma~\ref{LemmaA5X}. Therefore, the proof is the same.

The proof of part (c) is similar to that of
Theorem~\ref{thm:compwithoutK}, except that the following lemma is used
instead of Lemma~\ref{LemmaA5X}.

\lem
\label{LemmaA5K}
If $\varphi$ is a $\axKnrr\union\{\rm{T},4,5^*\}$-consistent sentence in $\LKn(\Phi)$, then $\varphi$ is satisfiable
in $\N_n^{r,t,e}(\Phi)$.
\elem

\prf Let $M^{c}= (\Sigma,\L, {\cal K}_1,...,{\cal
K}_n,\A_1,\ldots,\A_n,\pi)$ be a canonical extended awareness structure
constructed as follows
\begin{itemize}
\item $\Sigma=\{(s_V,L):V$ is
a set of sentences in $\LKn(L)$ that is maximal $\axKnrr\union\{\rm{T},4,5^*\}$-consistent with respect to $L$
and $L\subseteq\Phi$%
\};

\item $\L((s_V,L))=L$;

\item $
\pi((s_V,L),p)= \left\{
\begin{array}{ll}
                                {\bf true} & \mbox{if $p \in V$}, \\
                                {\bf false} & \mbox{if $p\in (L-V)$}; \\
                                \end{array}
                            \right.
                            $

\item $\A_i((s_V,L))$ is arbitrary;

\item ${\cal K}_i((s_V,L))= \{(s_W,L):V/K_i\subseteq W\}$.

\end{itemize}

It is easy to see that $M^c\in \N_n^{r,t,e}(\Phi)$. As usual, to prove Lemma~\ref{LemmaA5K}, we now show that for every $\psi\in \LKn(L)$,
\begin{eqnarray}
\label{eq:consiffsatis2} (M^{c},(s_V,L))\sat\psi\mbox{\ \ iff\ \
}\psi\in V.
\end{eqnarray}

We prove (\ref{eq:consiffsatis2}) by induction on the length of the
formula.
All the cases are standard, except for the case that $\psi=K_i\psi'$.
In this case,
if $\psi\in V$, then $\psi'\in W$ for every $W$
such that $(s_W,L')\in \K_i((s_V,L))$. By the induction hypothesis,
$(M^c,(s_W,L'))\sat \psi'$ for every $(s_W,L')\in \K_i((s_V,L))$, so
$(M^c,(s_V,L))\sat K_i\psi'$.

If $\psi\notin V$,
since $\psi\in\LKn(L)$, it follows that
$\neg\psi\in V$.
If $A_i^*\psi'\notin V$,
then $\psi'$ is not defined at some world $(s_W,L')\in\K_i((s_V,L))$ which implies that $(M^c,(s_V,L))\not\sat
\psi$.
If $A_i^*\psi'\in V$, then we need to show that $V/K_i\cup\{\neg\psi'\}$
is $\axKnrr\union\{\rm{T},4,5^*\}$-consistent. Suppose not. Then there
exists a subset $\{\beta_1,\ldots,\beta_k\}\subseteq V/K_i$ such that
$$\axKnrr\union\{\rm{T},4,5^*\}\vdash \beta\rimp \psi',$$
where $\beta=\beta_1\land\cdots\land\beta_k$. By Gen$^*$, it follows that
$$\axKnrr\union\{\rm{T},4,5^*\}\vdash A_i^*(\beta\rimp \psi')\rimp K_i(\beta\rimp \psi').$$
Since $\{\beta_1,\ldots,\beta_k\}\subseteq V/K_i$, it follows that $\{K_i\beta_1,\ldots,K_i\beta_k\}\subseteq V$.
Thus, by A0$^*$, we have
$\{A_i^*\beta_1,\ldots,A_i^*\beta_k\}\subseteq V$. Thus, $A_i^*(\beta\rimp \psi')\in V$ and $K_i\beta \in V$. Therefore, $K_i\psi'\in V$, a contradiction.

Since $V/K_i\cup\{\neg\psi'\}\subseteq \LKn(L)$ and is $\axKnrr\union\{\rm{T},4,5^*\}$-consistent, it follows that there exists a set of sentences $W$ that is maximal $\axKnrr\union\{\rm{T},4,5^*\}$-consistent with respect to $L$ and contains $V/K_i\cup\{\neg\psi'\}$. Thus, $(s_W,L)\in \K_i((s_V,L))$ and, by the induction hypothesis,
$(M^c,(s_W,L))\not\sat \psi'$. Thus, $(M^c,(s_V,L))\not\sat \psi$.
\eprf

\eprf
}
\subsubsection*{Acknowledgments}
The first author is supported in part by NSF grants ITR-0325453,
IIS-0534064, and IIS-0812045, and by AFOSR grants
FA9550-08-1-0438 and FA9550-05-1-0055.
The second author is supported in part by FACEPE under
grants APQ-0150-1.02/06 and APQ-0219-3.08/08, and by  MCT/CNPq under
grant 475634/2007-1.

\bibliographystyle{chicagor}
\bibliography{z,joe}
\end{document}